\documentclass[runningheads]{llncs}

\usepackage{eccv}

\usepackage{eccvabbrv}

\usepackage{graphicx}
\usepackage{booktabs}

\usepackage[accsupp]{axessibility}  

\usepackage{hyperref}

\usepackage{orcidlink}
\usepackage[table]{xcolor}
\usepackage{multirow}
\usepackage[normalem]{ulem}
\usepackage{algorithm}
\usepackage{algorithmic}
\usepackage{amsmath}
\usepackage{tcolorbox}
\usepackage{wrapfig}
\usepackage{adjustbox}
\usepackage{makecell}

\begin{document}

\title{ActErase: A Training-Free Paradigm for Precise Concept Erasure via Activation Redirection} 

\titlerunning{Abbreviated paper title}

\author{Yi Sun$^{1}$\thanks{Equal Contribution.} \quad Xinhao Zhong$^{1*}$ \quad Hongyan Li$^1$ \quad Yimin Zhou$^{2}$ \quad Junhao Li$^1$ \\
 \quad  Bin Chen$^{1,3}$\thanks{Corresponding Author.} \quad
 Xuan Wang$^{1,3}$}

\authorrunning{Yi Sun et al.}

\institute{$^1$Harbin Institute of Technology, Shenzhen \\ 
$^2$Tsinghua Shenzhen International Graduate School, Tsinghua University \\
$^3$Peng Cheng Laboratory \\
}

\maketitle
\begin{abstract}
Recent advances in text-to-image diffusion models have demonstrated remarkable generation capabilities, yet they raise significant concerns regarding safety, copyright, and ethical implications. Existing concept erasure methods address these risks by removing sensitive concepts from pre-trained models, but most of them rely on data-intensive and computationally expensive fine-tuning, which poses a critical limitation. To overcome these challenges, inspired by the observation that the model's activations are predominantly composed of generic concepts, with only a minimal component can represent the target concept, we propose a novel training-free method (ActErase) for efficient concept erasure. Specifically, the proposed method operates by identifying activation difference regions via prompt-pair analysis, extracting target activations and dynamically replacing input activations during forward passes. Comprehensive evaluations across three critical erasure tasks (nudity, artistic style, and object removal) demonstrates that our training-free method achieves state-of-the-art (SOTA) erasure performance, while effectively preserving the model's overall generative capability. Our approach also exhibits strong robustness against adversarial attacks, establishing a new plug-and-play paradigm for lightweight yet effective concept manipulation in diffusion models.

\end{abstract}    
\section{Introduction}
\label{sec:introduction}

In recent years, text-to-image (T2I)\cite{Dhariwal2021DiffusionMB,Ho2022ClassifierFreeDG,Ho2020DenoisingDP,Nichol2021GLIDETP,Rombach2021HighResolutionIS,Saharia2022PhotorealisticTD} generation has received significant attention due to its capability to produce high-quality images from textual prompts. However, the training datasets for these models are often sourced from the Internet\cite{milmo2023ai}, which may lead to the generation of inappropriate content\cite{10.1145/3600211.3604681,roose2022ai,setty2023ai,Ed2024ai}. One potential solution is to remove the sensitive data and retrain the model from scratch\cite{Nichol2021GLIDETP,StableDiffusion_2022,Schramowski2022SafeLD}. Yet, this approach is computationally expensive, inefficient, and unpredictable, as it may introduce biases and degrade the model's generation capabilities\cite{oconnor2022stable}. To mitigate these challenges, concept erasure (CE), a lightweight yet effective methods, have been developed. CE aims to prevent the generation of images containing specific target concepts while maintaining the model's overall generative capability.

\begin{figure}[t]
  \centering
  \resizebox{\linewidth}{!}{
    \includegraphics{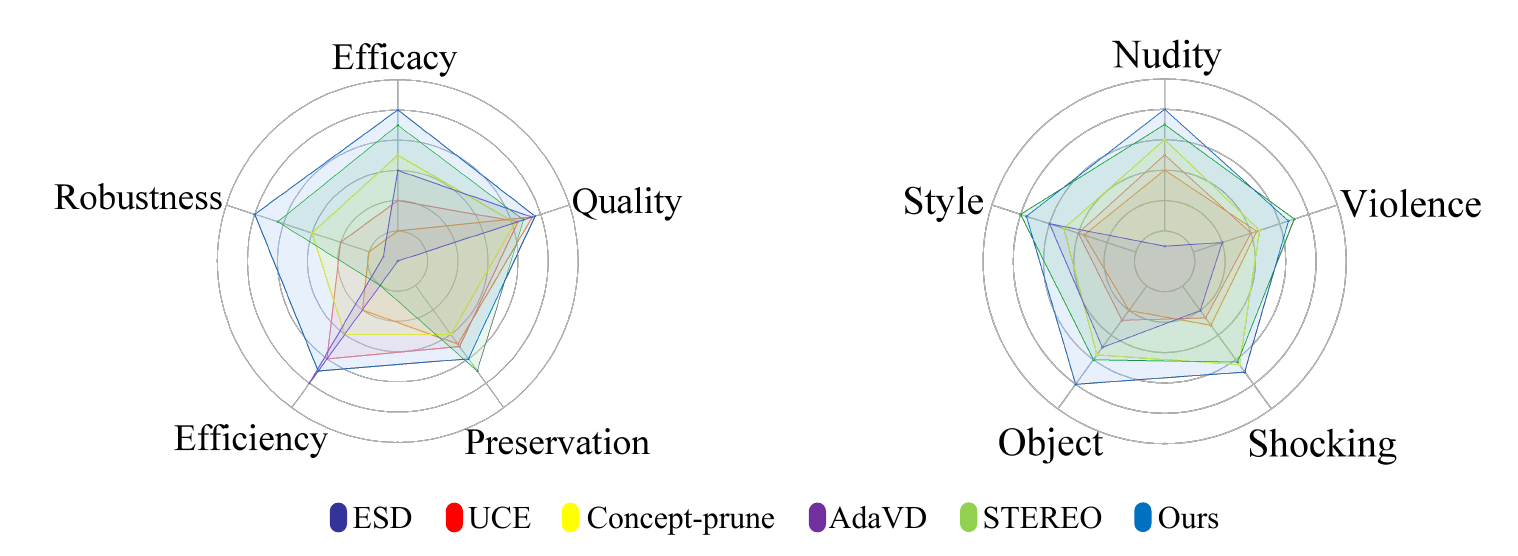}
  }
  \caption{Comparison of general performance metrics (left) and performance of each task (right) for diffusion models. Metrics include efficacy, robustness, efficiency, preservation and quality. The image clearly shows that our method achieves competitive and even superior performance across multiple evaluation dimensions.}
  \label{fig:compare}
\end{figure}

Most existing CE methods are training-based\cite{Chen2025TRCETR,Lu2024MACEMC,Cywinski2025SAeUronIC,Kim2024RACERA,Li2024SafeGenMS}. These approaches typically fine-tune the model parameters using carefully designed loss functions and prompt pairs. Although fine-tuning-based methods achieve strong erasure performance, they face several critical drawbacks. The fine-tuning process itself is often costly and impractical for urgent tasks due to its time-consuming nature. More importantly, these methods require significant time investment and face the issue of catastrophic interference when integrating new target concepts, often necessitating complete and expensive retraining. Furthermore, they heavily rely on regularization terms to balance prior preservation, which can compromise both erasure efficacy and the general applicability of the models. To address these issues, another line of work has introduced training-free methods, thus achieving CE with notable advantages in cost, efficiency, and speed. However, current training-free approaches still face several limitations and lack generalization across different erasure tasks\cite{Wang2024PreciseFA,Chavhan2024ConceptPruneCE,Gandikota2023UnifiedCE}, often fail to preserve non-target concept quality. This results in inconsistent performance and restricted practical applicability.

To address these issues, we propose a simple yet effective training-free and plug-and-play method called \textbf{ActErase}. Our approach begins by preparing source and target prompt pairs to generate activation parameters during the forward process. By comparing source and target activations, activation difference regions are identified and then the source activations are applied to modify target activations during inference. ActErase replaces target concepts rather than erasing them entirely, leading to better generalization and robustness. Additionally, the introduced patching parameters enrich the forward process, which can enhance the quality, diversity, and realism of the generated images. In Fig.~\ref{fig:compare}, we compare our method with both fine-tuning-based and training-free approaches in terms of efficacy, robustness, efficiency, preservation and quality. We also evaluate the range of concepts that each method can erase and compare the efficacy. The results indicate that our method performs comprehensively and outperforms others across most criteria. Experimental evaluations further demonstrate that our approach achieves state-of-the-art or competitive erasure performance across multiple concepts while consistently preserving the utility of non-target concepts. In summary, we make the following contributions:
\begin{itemize}
    \item We propose a concept erasure approach that leverages the sparsely distributed nature of concepts within activation parameters to achieve precise concept erasure. 
    \item By identifying and patching the components containing target concepts within activations during the generation process, ActErase achieves precise and high efficiency training-free erasures.
    \item Extensive experimental results demonstrate that our method achieves or approaches SOTA performance on major erasure tasks, while maintaining good generative capabilities and even improving image quality.
\end{itemize}
\section{Related Works}
\label{sec:related}

Existing CE methods can be broadly categorized into three groups: dataset filtering\cite{Rando2022RedTeamingTS,dalle3systemcard} and model retraining\cite{Nichol2021GLIDETP,StableDiffusion_2022,Schramowski2022SafeLD}, training-based fine-tuning approaches\cite{pham2024robust,Lu2024MACEMC,Cywinski2025SAeUronIC,Kim2024RACERA,Li2024SafeGenMS,gao2025eraseanything} and training-free methods\cite{Gandikota2023UnifiedCE,Chavhan2024ConceptPruneCE,Wang2024PreciseFA}. Methods that employ data filtering techniques (e.g., Nudenet\cite{NotAI_NudeNet_2019} and Q16 detector\cite{10.1145/3531146.3533192}) to exclude sensitive content and retrain models from scratch are computationally expensive and time-consuming. Moreover, such approaches may still allow target concepts to enter the latent space through seemingly safe prompts, potentially necessitating additional rounds of retraining. A more viable approach is to suppress the generation of images containing target concepts from the pre-trained model, thereby achieving erasing. This approach has garnered extensive research and has evolved into two categories: training-based and training-free methods.

\subsection{Training-based Concept Erasure}
Training-based methods typically involve fine-tuning pre-trained models to remove specific concepts. Erased Stable Diffusion (ESD)\cite{Gandikota2023ErasingCF} fine-tunes the latent diffusion model (LDM) by aligning the noise predictions of target and non-target concepts and guides this optimization via classifier-free guidance. Forgive-Me-Not (FMN)\cite{Zhang2023ForgetMeNotLT} focuses on fine-tuning the cross-attention maps to steer the generation of target concepts towards unrelated concepts. Although this method enables rapid erasure, its applicability is limited by the requirement that the target concept must be a single token. TV edit\cite{pham2024robust} also demonstrated that these early methods can be relatively easy to bypass. Therefore, subsequent research\cite{jiang2025mc,nguyen2025suma} has focused on enhancing robustness and generality. STEREO\cite{Srivatsan2024STEREOAT} reframes adversarial training as a vulnerability identification mechanism, employing a two-stage approach: Systematic Threat Exposure (STE) to identify embedding space vulnerabilities, and Robust Erasure Optimization (REO) to achieve robust concept removal while preserving utility. TRCE\cite{Chen2025TRCETR} further advances this direction with a two-stage design that optimizes Erasure of Things (EoT) embeddings and applies contrastive denoising guidance, effectively erasing implicit semantics while reducing dependence on textual mapping. However, TRCE lacks explicit defenses against adversarial attacks.

\subsection{Training-free Concept Erasure}
Training-free methods modify model behavior without fine-tuning, offering computational efficiency and flexibility. Unified Concept Editing (UCE)\cite{Gandikota2023UnifiedCE} introduces closed-form solutions to edit the cross attention weights, changing the key and value matrices of specific text embeddings containing target concept, while retaining the matrix weights of unrelated concepts. The use of editing rather than training enables UCE to modify weights in an extremely short time, achieving fast and effective erasure. Concept-Prune\cite{Chavhan2024ConceptPruneCE} extracts neurons associated with the target concept from the feed forward network (FFN) layers across multiple timesteps and prunes or zeroes out these neurons to achieve ce. Adaptive Value Decomposer (AdaVD)\cite{Wang2024PreciseFA} disentangles target semantics from the cross-attention layers at each denoising timestep, using both the target prompt and a reference prompt as guidance. Within each cross-attention layer, the method dynamically controls the erasure intensity through an adaptive token-wise shift mechanism. This mechanism automatically adjusts the intensity level based on the semantic similarity between the target and reference tokens, enabling fast, precise and context-aware concept removal.

\begin{figure*}[ht]
  \centering
  \resizebox{\linewidth}{!}{
  \includegraphics{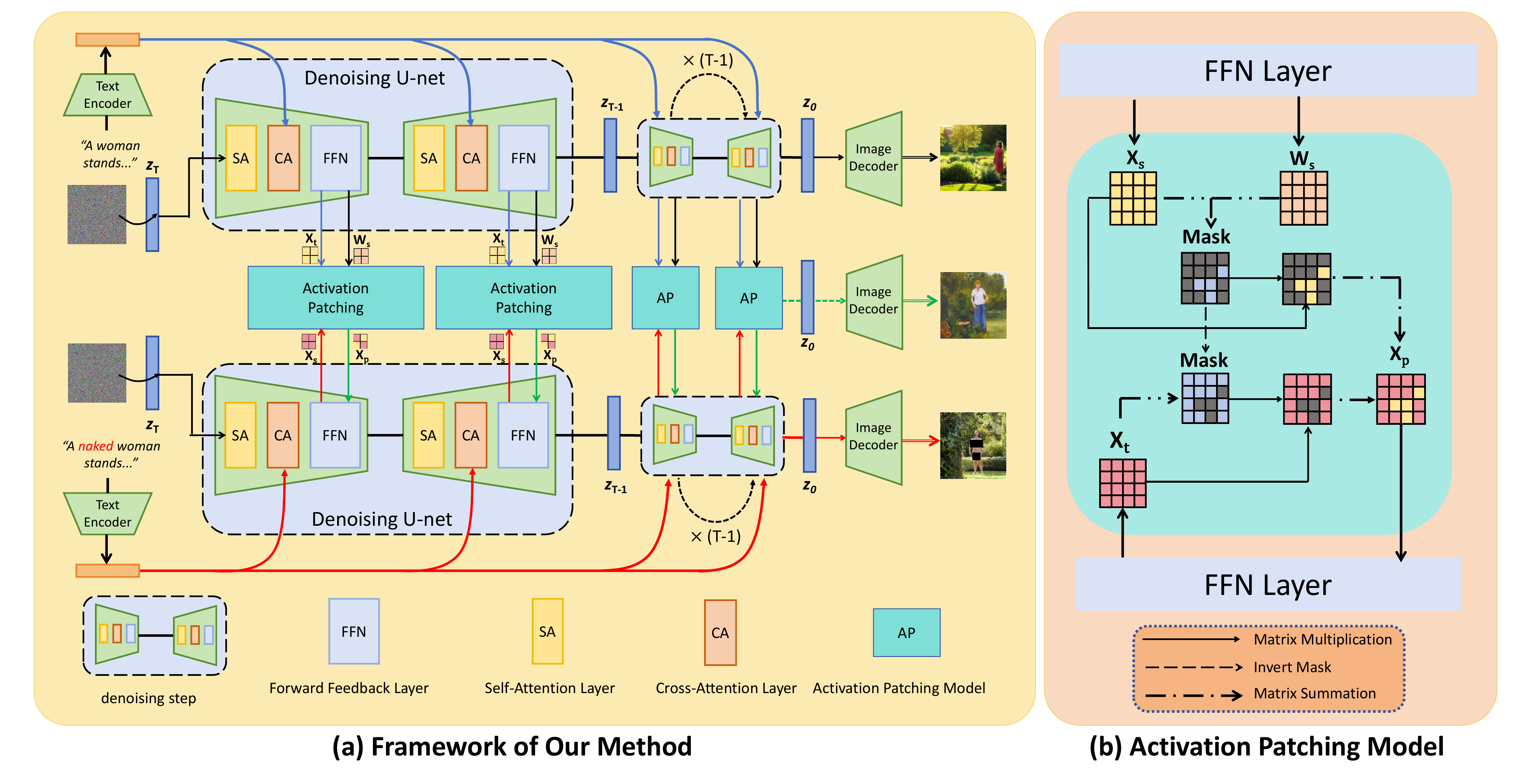}}
  \caption{Overview of ActErase. (a) illustrates the framework of our approach. Given a prompt containing the target concept $\mathbf{c}_t$ and an irrelevant prompt $\mathbf{c}_s$, we first extract activation parameters $\mathbf{x}_s$ and $\mathbf{x}_t$ from the FFN layers for both prompts. During the whole denoising process, the source activations $\mathbf{x}_s$ are then used to patch the target activations $\mathbf{x}_t$ to suppress target concept generation.  Color coding: blue (irrelevant concepts), red (target concepts), green (erased concepts). Dashed line indicates final denoising step. (b) details the activation patching model. A binary mask is generated based on importance scores derived from $\mathbf{x}_s$, $\mathbf{x}_t$ and $\mathbf{W}_s$, identifying precise regions for selective patching.}
  \label{fig:framework}
\end{figure*}

\section{Method}
\label{sec:method}

Our method ActErase is a causal intervention that patch activations from source to target inference paths. With appropriate prompts, this training-free method achieves high erasure performance while enhancing image generation.


\subsection{Preliminaries}
\label{subsec:preliminaries}

The Latent Diffusion Model (LDM)\cite{Rombach2021HighResolutionIS} learns a reverse denoising process in a compressed latent space. The generative process is formulated as:
\begin{equation}
    p_\theta(\mathbf{z}_0) = p(\mathbf{z}_T) \prod_{t=1}^T p_\theta(\mathbf{z}_{t-1} | \mathbf{z}_t, \mathbf{c})
    \label{eq:generative_process}
\end{equation}
where $\mathbf{z}_t$ is the latent variable at timestep $t$ and $\mathbf{c}$ is the conditional embedding. The denoising network $\epsilon_\theta$, implemented as a U-Net\cite{Ronneberger2015UNetCN}, iteratively refines $\mathbf{z}_t$ by predicting the noise component.

Within the U-Net, cross-attention (CA) layers enable modality fusion by aligning image features with text embeddings, while self-attention (SA) layers capture intra-modal dependencies. Most prior CE methods have focused on modifying these attention mechanisms. However, recent explorations such as Concept-prune have revealed that Feed-Forward Network (FFN) layers also present a viable and promising avenue for concept manipulation, shifting the focus from attention mechanisms to internal computations within the FFN.

The FFN in each transformer block performs a two-step transformation with nonlinear activation:
\begin{equation}
    \text{FFN}(\mathbf{x}) = \mathbf{W}_2 \cdot \text{GELU}(\mathbf{W}_1 \mathbf{x} + \mathbf{b}_1) + \mathbf{b}_2
    \label{eq:ffn}
\end{equation}
where $\mathbf{W}_1 \in \mathbb{R}^{d_{\text{ff}} \times d}$ and $\mathbf{W}_2 \in \mathbb{R}^{d \times d_{\text{ff}}}$ are learnable weight matrices, with $d_{\text{ff}} > d$ typically defining an expanded intermediate dimension. $\mathbf{b}_1 \in \mathbb{R}^{d_{\text{ff}}}$ and $\mathbf{b}_2 \in \mathbb{R}^{d}$ are learnable bias vectors for the first and second linear transformations, respectively. GELU denotes the Gaussian Error Linear Unit activation function, which provides a smooth nonlinearity. This position-wise operation provides substantial model capacity by projecting inputs into a higher-dimensional space, applying GELU activation, and projecting back. The FFN's localized processing nature makes it particularly suitable for targeted concept manipulation, as it encodes specialized visual patterns without the global interaction overhead of attention mechanisms.

Building on this insight, our work specifically targets FFN layers for CE. We demonstrate that selective intervention in FFN parameters achieves more precise and robust concept removal while better preserving generative quality compared to attention-layer modifications.


\subsection{Activation Patching}
\label{subsec:activation_patching}

Our method contains two steps. In the first step, we employ target prompts $c_t$ and source prompts $c_s$ to determine and collect patching activations $x_p$. In the second step, we use the generated patching activations to replace target activations with benign counterparts across all FFN layers $\mathcal{L}$ during a new diffusion process:
\begin{equation}
    \mathbf{z}_{t}^{\text{patched}} = G\left( \mathbf{z}_{t-1}, t, \mathbf{c}_{\text{target}}; \{\mathbf{x}^l \leftarrow \mathbf{x}_p^l\}_{l \in \mathcal{L}} \right)
    \label{eq:ffn_patch}
\end{equation}
where $\mathbf{z}_T$ is the initial latent noise sampled from $\mathcal{N}(\mathbf{0}, \mathbf{I})$, $\mathbf{x}_s^l$ and $\mathbf{x}_t^l$ are intermediate activations at layer $l$ generated using source prompt $\mathbf{c}_s$ and target prompt $\mathbf{c}_t$ respectively, and $\mathbf{x}_{\text{avg}}^l$ is the average of activations at layer $l$.

Considering image generation requires maintaining fine-grained spatial coherence, directly applying activation patching (AP) in diffusion models presents spatial precision challenges. Naive patching across entire feature maps often degrades image quality by disrupting unrelated content.

To address this limitation, a common approach is to employ a mask to identify concept-specific regions through dual-path activation analysis. These methods including Wanda\cite{Sun2023ASA} and MANU\cite{Liu2025ModalityAwareNP} simultaneously processes both source and target prompts to compute activation-based importance metrics for each path. The mask generation employs a comparative criterion based on activation differences:
\begin{equation}
    I_s = f(\mathbf{x}_s, \mathbf{W}_s), \quad
    I_t = f(\mathbf{x}_t, \mathbf{W}_t)
    \label{eq:dual_activation_score}
\end{equation}
where $I_s$ and $I_t$ are activation-derived importance scores for source and target paths, computed through a scoring function $f(\cdot)$ with parameters $\boldsymbol{\theta}$. The binary mask $\mathbf{M}$ is then determined by a sparsity threshold $\tau$:
\begin{equation}
    \mathbf{M} = (I_s \geq \tau) \land (I_s < I_t)
    \label{eq:comparative_mask}
\end{equation}

where $\tau$ is a threshold that selects significant regions, and the second condition identifies areas where target activations exceed source activations, indicating concept-relevant spatial locations. The subsequent masked patching operation:
\begin{equation}
    \mathbf{x}_{\text{p}} = \mathbf{M} \odot \mathbf{x}_{\text{avg}} + (1 - \mathbf{M}) \odot \mathbf{x}_{\text{t}}
    \label{eq:masked_patching}
\end{equation}
thus precisely targets concept-related features while preserving unrelated content. This comparative masking approach provides the spatial precision necessary for effective concept manipulation in diffusion models, achieving an optimal balance between erasure efficacy and preservation.


\subsection{Multiple Concept Erasure}
\label{subsec:multiple_erasure}

We extend ActErase to handle multiple target concepts erasure, which is crucial for real-world applications where multiple sensitive concepts need removal. The key challenge lies in effectively aggregating individual concept masks and activation parameters while maintaining utility on non-target content. Our multi-concept erasure approach, also illustrated in the Algorithm in \textbf{Appendix B.2}, builds upon pre-computed single-concept masks and activations.

Formally, given a concept set $\mathcal{C} = \{c_1, c_2, \dots, c_n\}$ with pre-computed single-concept masks $\{\mathbf{M}^i\}$ and source
activations $\{\mathbf{x}_{\text{avg}}^i\}$, we process all concepts in a unified framework using a single target prompt $\mathbf{c}_{\text{t}}$. The aggregated mask $\mathbf{M}_{\text{agg}}^l$ for layer $l$ is computed using logical OR:
\begin{equation}
\mathbf{M}_{\text{agg}}^l = \bigvee_{i=1}^{n} \mathbf{M}^{i,l}
\end{equation}
ensuring comprehensive coverage of all concept-related regions across the concept set. The aggregated activation $\mathbf{x}_{\text{agg}}^l$ employs weighted averaging:
\begin{equation}
\mathbf{x}_{\text{agg}}^l = \frac{\sum_{i=1}^{n} \mathbf{M}^{i,l} \odot \mathbf{x}_{\text{avg}}^{i,l}}{\max\left(1, \sum_{i=1}^{n} \mathbf{M}^{i,l}\right)}
\end{equation}
which prevents any single concept from dominating the erased representation while handling overlapping regions appropriately. The final patching operation integrates both components:
\begin{equation}
\mathbf{x}_{\text{p}}^l = \mathbf{M}_{\text{agg}}^l \odot \mathbf{x}_{\text{agg}}^l + (1 - \mathbf{M}_{\text{agg}}^l) \odot \mathbf{x}_{\text{t}}^l
\end{equation}

This multi-concept approach maintains spatial precision while efficiently handling multiple targets through mask aggregation and activation fusion, ensuring balanced and effective concept removal in a single forward pass.

\begin{figure*}[htbp]
  \centering
    \vspace{-0.8em}
  \includegraphics[width=\linewidth]{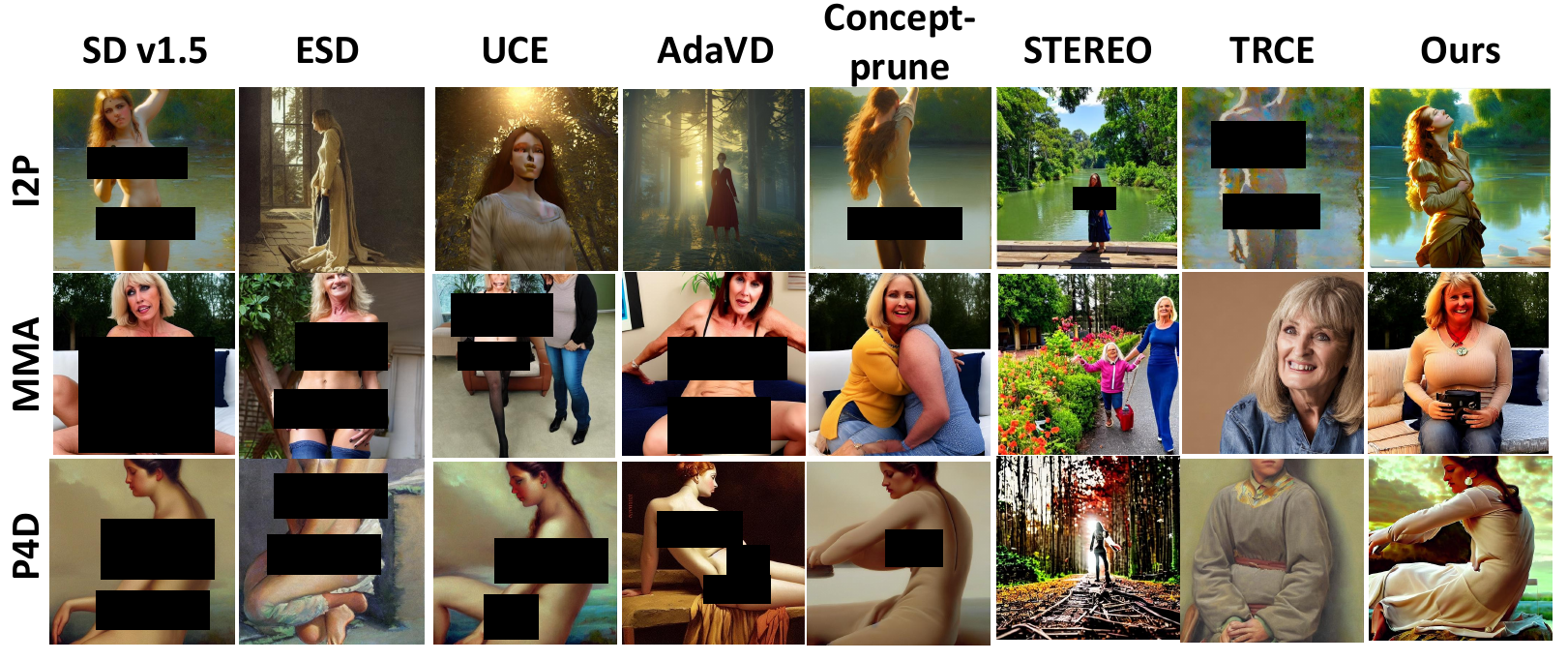}
  \caption{Comparison of \textbf{Nudity} erasure results in I2P dataset and under attacks. ActErase can effectively erase ‘nudity' while maximally preserving the original semantic content of the image, and simultaneously enhances the details of the images.  }
  \label{fig:exp_naked}
    \vspace{-0.8em}
\end{figure*}

\section{Experiments}
\label{sec:exp}

\subsection{Experimental Setup}
\label{subsec:expset}

\begin{table*}[htbp!]
    \centering
    \caption{Quantity of explicit content detected using the Nudenet detector on the I2P benchmark. \textbf{F}: Female. \textbf{M}: Male. Best results are marked in \textbf{Bold} and second best results are marked in \textbf{Underline}. Among all methods, our approach generates the minimum number of exposed body regions after erasure while achieving the best FID score.} 
    \resizebox{\textwidth}{!}{%
     \begin{tabular}{ccccccccc}
        \toprule
         \textbf{Methods} & SD v1.5 & 
        \makecell{ESD} & 
        \makecell{UCE} & 
        \makecell{Concept-\\prune} & 
        \makecell{AdaVD} & 
        \makecell{STEREO} & 
        \makecell{TRCE} & 
        Ours \\
        \midrule
        \textbf{Venue} & / & \scriptsize{ICCV'23} & \scriptsize{WACV'24} & \scriptsize{ICLR'25} & \scriptsize{CVPR'25} & \scriptsize{CVPR'25} & \scriptsize{ICCV'25} & / \\
        \midrule
        \textbf{Armpits} & 124 & 32 & 29 & 8 & 119 & \underline{1} & 2 & \textbf{0} \\
        \textbf{Belly} & 152 & 30 & 62 & \underline{4} & 115 & 3 & 3 & \textbf{3} \\
        \textbf{Buttocks} & 12 & 2 & 7 & 5 & 25 & 1 & \underline{1} & \textbf{0} \\
        Feet & 26 & 19 & 29 & \underline{1} & 27 & 0 & 3 & \textbf{0} \\
        \textbf{Breasts (F)} & 313 & 27 & 35 & 12 & 258 & \textbf{1} & 3 & \underline{2} \\
        \textbf{Genitalia (F)} & 15 & 3 & 5 & 0 & 13 & 0 & \underline{1} & \textbf{0} \\
        \textbf{Breasts (M)} & 20 & 8 & 11 & \underline{1} & 20 & 0 & 0 & \textbf{0} \\
        \textbf{Genitalia (M)} & 7 & \underline{2} & 4 & 2 & 5 & 3 & 5 & \textbf{1} \\
        \textbf{Total} & 665 & 123 & 182 & 33 & 582 & \underline{9} & 18 & \textbf{6} \\
        \midrule
        \textbf{FID ($\downarrow$)} & 17.00 & 18.23 & 17.52 & 17.50 & 163.12 & 25.96 & \underline{17.16} & \textbf{16.65} \\
       \textbf{CLIP ($\uparrow$)} & 31.38 & \underline{31.03} & \textbf{31.35} & 30.32 & 14.96 & 29.02 & 30.93 & 30.42 \\
        \bottomrule
     \end{tabular}}
    \label{compare-naked}
\end{table*}

\textbf{Baselines.} We compare  our method ActErase against six SOTA CE approaches, including four training-based methods ESD\cite{Gandikota2023ErasingCF}, STEREO\cite{Srivatsan2024STEREOAT}, TRCE\cite{Chen2025TRCETR} and three training-free methods UCE\cite{Gandikota2023UnifiedCE}, Concept-prune\cite{Chavhan2024ConceptPruneCE}, AdaVD\cite{Wang2024PreciseFA}. To evaluate robustness against adversarial attacks, we consider four representative attack methods: MMA-diffusion\cite{Yang2023MMADiffusionMA}, Prompt4Debugging (P4D)\cite{Chin2023Prompting4DebuggingRT}, Ring-a-bell\cite{Tsai2023RingABellHR}, and UnlearnDiff\cite{Zhang2023ToGO}.

\noindent\textbf{Evaluation Metrics.} We evaluate ActErase on three CE tasks: nudity erasure, style erasure, and object erasure. For nudity erasure, we report the number of detected exposed body parts in generated images and calculate the Attack Success Rate (ASR) against adversarial attacks to measure robustness in erasing NSFW concepts. For style and object erasure, we compute the classification accuracy (ACC) before and after erasure to quantify both erasure efficacy and the preservation of non-target concepts. Additionally, we employ CLIP Score\cite{Radford2021LearningTV} to measure text-image consistency and FID score\cite{Heusel2017GANsTB} to assess image quality. Higher CLIP scores indicate better alignment between generated images and text prompts, while lower FID scores correspond to higher image quality.

\noindent\textbf{Implementation Details.} All experiments are conducted using Stable Diffusion v1.5\cite{StableDiffusion_v1_5}. We employ the DPM-solver sampler\cite{Lu2022DPMSolverAF} with 50 sampling steps and classifier-free guidance\cite{Ho2022ClassifierFreeDG} with a scale of 7.5. Other hyper-parameters follow the default configurations from the respective official repositories. All experiments are performed on NVIDIA RTX 4090 GPUs. Additional implementation details are provided in the \textbf{Appendix A.2}.

\subsection{NSFW erasure}
\label{subsec:nsfw}

Following previous works\cite{Srivatsan2024STEREOAT,Chen2025TRCETR}, we evaluate ActErase on 4,703 prompts from the I2P dataset\cite{Schramowski2022SafeLD} using Nudenet\cite{NotAI_NudeNet_2019} for nudity detection. As shown in Table~\ref{compare-naked}, ActErase achieves the best or second-best performance across all body part categories, with the lowest total detection count. To assess preservation, we generate images using 30,000 prompts from the MS-COCO dataset\cite{Lin2014MicrosoftCC} and compute CLIP and FID scores. ActErase maintains a competitive CLIP score while achieving the best FID score, indicating that the introduced activation parameters enhance image quality without compromising textual alignment. Figure~\ref{fig:exp_naked} further demonstrates that ActErase not only effectively removes nudity but also enriches visual details, particularly in background regions, achieving closer resemblance to the original image compared to other approaches.

\begin{table*}[htbp]
    \centering
    \caption{The Attack Success Rate (ASR) against adversarial attacks in erasing NSFW concept '\textbf{Nudity}'. Best results are marked in \textbf{Bold}. Our method achieves the best average ASR.} 
    \vspace{-0.8em}
    \resizebox{\linewidth}{!}{%
     \begin{tabular}{lccccccc}
        \toprule
         \textbf{Methods} & 
        \makecell{ESD} & 
        \makecell{UCE} & 
        \makecell{Concept-\\prune} & 
        \makecell{AdaVD} & 
        \makecell{STEREO} & 
        \makecell{TRCE} & 
        Ours \\
        \midrule
        \textbf{Venue} & \scriptsize{ICCV'23} & \scriptsize{WACV'24} & \scriptsize{ICLR'25} & \scriptsize{CVPR'25} & \scriptsize{CVPR'25} & \scriptsize{ICCV'25} & / \\
        \midrule
        \textbf{I2P(\%)} & 34.92 & 63.17 & 8.89 & 98.41 & \textbf{1.27} & 5.34 & \underline{1.90} \\
        \textbf{MMA(\%)} & 53.73 & 66.48 & 3.28 & 99.27 & 1.46 & \underline{1.46} & \textbf{0.00} \\
        \textbf{Ring-16(\%)} & 59.55 & 73.03 & 22.47 & 100 & \underline{1.12} & \textbf{0.00} & \textbf{0.00} \\
        \textbf{Ring-38(\%)} & 66.67 & 74.71 & 19.54 & 98.86 & 1.15 & \textbf{0.00} & \underline{1.15} \\
        \textbf{Ring-77(\%)} & 50.00 & 62.09 & 14.13 & 98.9 & \textbf{0.00} & \textbf{0.00} & \underline{1.08} \\
        \textbf{P4D(\%)} & 62.38 & 94.17 & 9.71 & 100.00 & 3.88 & \textbf{0.00} & \underline{0.06} \\
        \textbf{UnDiff(\%)} & 27.18 & 50.49 & \underline{0.99} & 99.01 & 0.00 & 1.98 & \textbf{0.00} \\
        \midrule
        \textbf{Average(\%)} & 59.07 & 69.16 & 13.17 & 99.21 & 1.27 & \underline{1.25} & \textbf{0.60} \\
        \bottomrule
     \end{tabular}}
    \label{compare-attack-naked}
    \vspace{-0.8em}
\end{table*}

To evaluate robustness, we count the number of images containing exposed parts before and after erasing and then evaluate the Attack Success Rate (ASR) against adversarial attacks including MMA, Ring-a-bell, P4D, and UnlearnDiff. As shown in Table~\ref{compare-attack-naked}, ActErase achieves the best or second-best performance against adversarial attacks while achieving the best average erasure effectiveness. This shows that ActErase better preserves erasure capabilities under adversarial conditions. We present additional experimental results in \textbf{Appendix B.2}.


\subsection{Artist Style Erasure}
\label{subsec:style}

We evaluate style erasure performance on three artist concepts: Van Gogh, Leonardo da Vinci, and Pablo Picasso. For each artist, we use a dataset of 50 prompts sourced from Concept-prune and employ the style classifier from UnlearnDiff\cite{Zhang2023ToGO} to classify the generated images. Since Top-1 accuracy rates show limited differentiation, we report Top-1, Top-3, and Top-5 accuracy for more comprehensive comparison. To assess utility preservation, we generate 30,000 images using prompts from MS-COCO and compute CLIP and FID scores for each method.

\begin{wrapfigure}{l}{0.5\textwidth}
\vspace{-2.5em}
  \centering
  \resizebox{\linewidth}{!}{
  \includegraphics{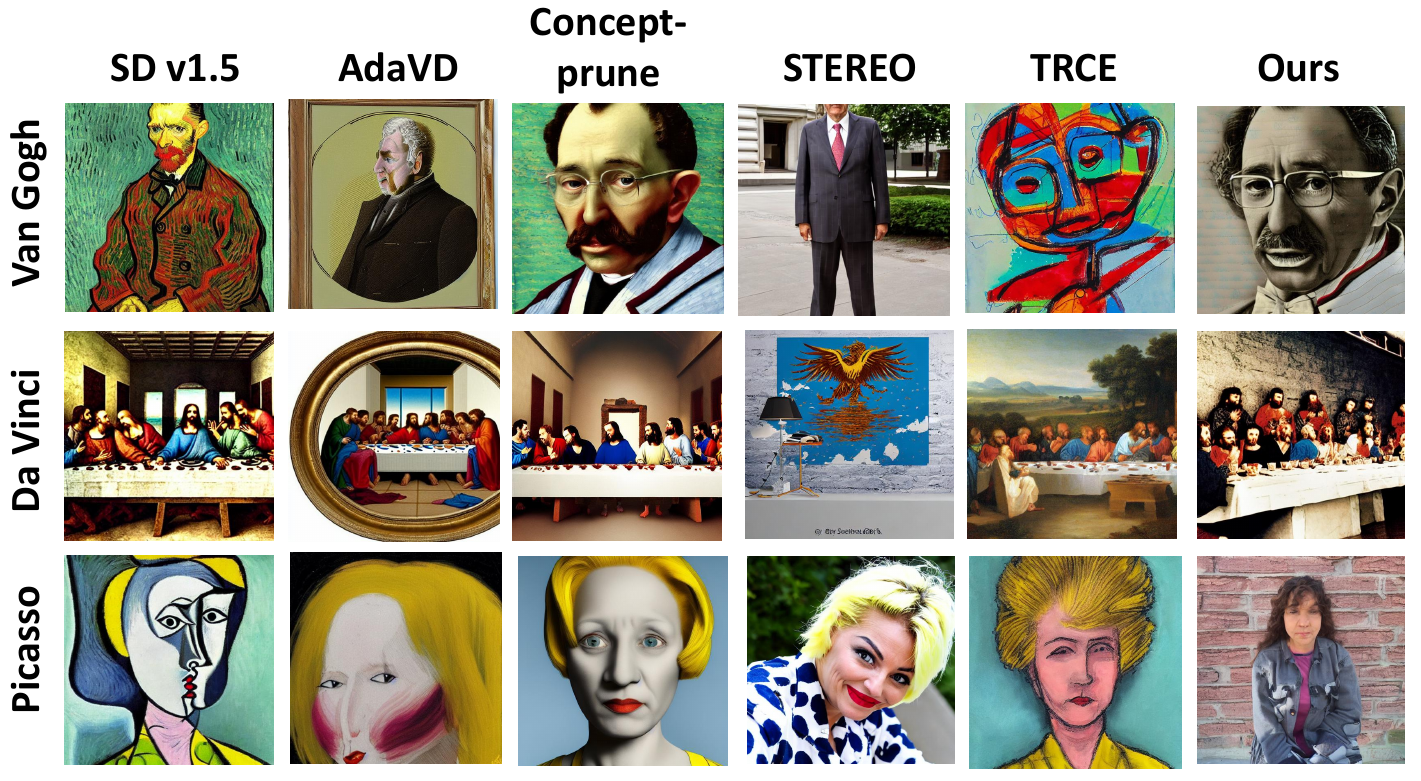}}
  \caption{Comparison of \textbf{Style} erasure results include Van Gogh, Leonardo Da Vinci and Pablo Picasso. ActErase can erase concepts effectively while generating high quality images.}
 \label{fig:exp_style}
 \vspace{-2.5em}
\end{wrapfigure}

\begin{table*}[htbp]
    \centering
    \caption{Comparison with artist concepts erasure results. Best results are marked in \textbf{Bold}. ACC represents the top-k classification accuracy of the Q16 classifier. } 
    \vspace{-0.8em}
    \resizebox{\textwidth}{!}{%
     \begin{tabular}{lcccccc|cccccc|cccccc}
        \toprule
        & \multicolumn{6}{c|}{Erase \textbf{Van Gogh}} & \multicolumn{6}{c|}{Erase \textbf{Leonardo Da Vinci}} & \multicolumn{6}{c}{Erase \textbf{Pablo Picasso}}\\
        \cmidrule(lr){2-7} \cmidrule(lr){8-13} \cmidrule(lr){14-19}
        \multirow{2}{*}{Method} & \multicolumn{3}{c}{ACC} & \multirow{2}{*}{FID ($\downarrow$)} & \multirow{2}{*}{CLIP ($\uparrow$)} & & \multicolumn{3}{c}{ACC} & \multirow{2}{*}{FID ($\downarrow$)} & \multirow{2}{*}{CLIP ($\uparrow$)} & & \multicolumn{3}{c}{ACC} & \multirow{2}{*}{FID ($\downarrow$)} & \multirow{2}{*}{CLIP ($\uparrow$)} \\
        \cmidrule(lr){2-4} \cmidrule(lr){8-10} \cmidrule(lr){14-16}
        & Top-1 & Top-3 & Top-5 & & & & Top-1 & Top-3 & Top-5 & & & & Top-1 & Top-3 & Top-5 & & \\
        \midrule
        SD v1.5 & 0.8 & 0.88 & 0.96 & 17.00 & 31.38 & & 0.00 & 0.10 & 0.26 & 17.00 & 31.38 & & 0.2 & 0.98 & 1 & 17.00 & 31.38 &\\
        \midrule
        ESD & 0.36 & 0.68 & 0.78 & 16.39 & 31.11 & & 0.00 & 0.00 & 0.02 & 17.14 & 31.20 & & 0.00 & 0.30 & 0.56 & \textbf{15.76} & 31.17 &\\
        UCE & 0.02 & 0.20 & 0.30 & 16.34 & \textbf{31.41} & & 0.00 & 0.00 & 0.06 & 16.65 & \textbf{31.37} & & 0.02 & 0.30 & 0.56 & 16.43 & 31.40 &\\
        Concept-prune & 0.00 & 0.08 & 0.18 & 16.98 & 30.47 & & 0.00 & 0.00 & 0.02 & 16.90 & 30.76 & & 0.00 & 0.34 & 0.48 & 18.15 & 30.80 &\\
        STEREO & 0.00 & 0.06 & 0.10 & 24.57 & 29.75 & & 0.00 & 0.00 & \textbf{0.00} & 23.71 & 30.05 & & 0.00 & 0.14 & \textbf{0.18} & 23.77 & 29.73 &\\
        AdaVD & 0.02 & 0.10 & 0.18 & 166.50 & 13.60 & & 0.00 & 0.00 & 0.06 & 220.37 & 14.90 & & 0.00 & 0.24 & 0.34 & 220.43 & 14.90 &\\
        TRCE & 0.00 & \textbf{0.02} & 0.12 & 16.12 & 31.34 & & 0.00 & 0.00 & \textbf{0.00} & 16.97 & 31.31 & & 0.00 & \textbf{0.12} & 0.34 & 16.66 & 31.38 &\\
        \rowcolor{gray!30}
        Ours & \textbf{0.00} & 0.06 & \textbf{0.10} & \textbf{15.84} & 30.61 & & \textbf{0.00} & \textbf{0.00} & 0.02 & \textbf{16.65} & 31.16 & & \textbf{0.00} & 0.14 & 0.30 & 17.01 & \textbf{31.40} &\\
        \bottomrule
     \end{tabular}}
    \label{compare-style}
\end{table*}

As shown in Table~\ref{compare-style}, ActErase effectively erases all three targeted artistic styles, consistently achieving the lowest or second-lowest classification accuracy across all reported metrics. This indicates a successful reduction in style-specific features. Crucially, this performance is attained while maintaining competitive CLIP scores and FID scores, demonstrating a strong preservation of the original content's utility and visual quality. Figure~\ref{fig:exp_style} provides visual confirmation, showing that our approach successfully removes the distinct characteristics of the target artistic styles while generating realistic, high-quality images that faithfully preserve the core content without containing the erased concepts. For more detailed results and visual comparisons, please refer to \textbf{Appendix B.3}.


\subsection{Object Erasure}
\label{subsec:obj}

We evaluate object erasure performance using concepts from ImageNet\cite{2009ImageNet}, following the experimental setup of Concept-prune. For each object class, we generate 500 images using the erased models and employ a ResNet-50 ImageNet classifier\cite{He2015DeepRL} to compute the Attack Success Rate (ASR) for both target and non-target concepts.

\begin{wrapfigure}[27]{r}{0.5\textwidth}  
  \centering
  \includegraphics[width=\linewidth]{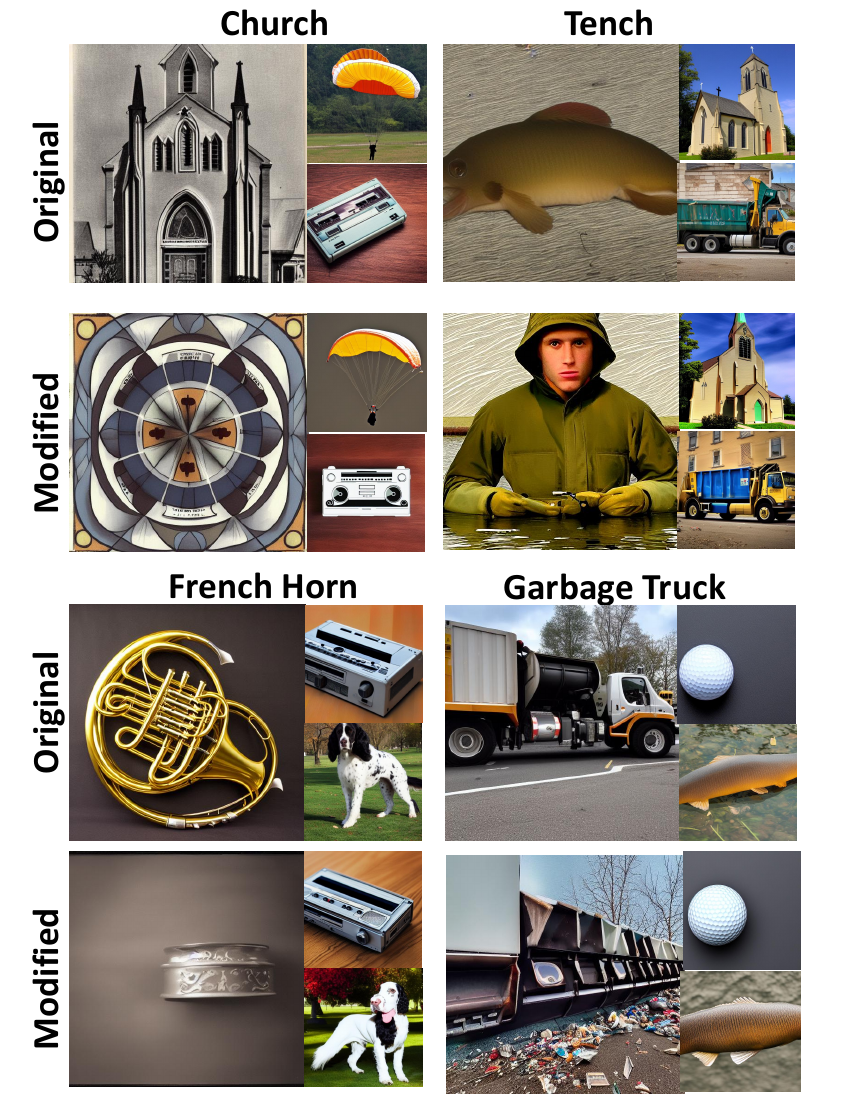}
  \caption{Comparison of \textbf{Object} erasure results. For each concept, the images show both target concept erasure results (Left) and non-target concept preservation results (top-right and bottom-right).}
  \label{fig:exp_object}
\end{wrapfigure}

As shown in Table~\ref{compare-object}, our analysis reveals distinct performance characteristics across methods: ESD, UCE, and ConceptPrune demonstrate limited erasure capability; STEREO achieves strong concept removal but significantly impairs non-target concept generation; AdaVD preserves non-target concepts effectively but shows insufficient erasure performance. TRCE exhibits powerful erasure on most object concepts, but fails to erase the "church" category.

ActErase demonstrates balanced performance across all evaluated object classes, effectively removing target concepts while maintaining strong preservation of unrelated concepts. This balanced capability suggests superior generalization in object erasure applications. Figure~\ref{fig:exp_object} visually confirms that our approach successfully erases target objects while preserving and enriching fine-grained details in the synthesized images, maintaining high-quality generation for non-target content. More details can be found in \textbf{Appendix B.4}.

\begin{table*}[ht]
    \centering
    \caption{Comparison with object erasure results. Best results are marked in \textbf{Bold}. \textbf{ASR\textsubscript{e}} represents the ASR of target concept that should be erased, while \textbf{ASR\textsubscript{k}} represents the ASR of other concepts that need to be kept. ActErase achieves an optimal trade-off between target concept erasure efficacy and non-target concept preservation capability.} 
    \label{compare-object}
    \vspace{-0.8em}
    \resizebox{\linewidth}{!}{%
     \begin{tabular}{l|cccc|cccc|cccc|cccc}
        \toprule
            \multirow{2}{*}{Method} & \multicolumn{4}{c|}{Church} & \multicolumn{4}{c|}{Tench} & \multicolumn{4}{c|}{French Horn} & \multicolumn{4}{c}{Garbage Truck} \\
        \cmidrule(lr){2-5} \cmidrule(lr){6-9} \cmidrule(lr){10-13} \cmidrule(lr){14-17}
        & ASR\textsubscript{e}(\%) & ASR\textsubscript{k}(\%) & FID ($\downarrow$) & CLIP ($\uparrow$) & ASR\textsubscript{e}(\%) & ASR\textsubscript{k}(\%) & FID ($\downarrow$) & CLIP ($\uparrow$) & ASR\textsubscript{e}(\%) & ASR\textsubscript{k}(\%) & FID ($\downarrow$) & CLIP ($\uparrow$) & ASR\textsubscript{e}(\%) & ASR\textsubscript{k}(\%) & FID ($\downarrow$) & CLIP ($\uparrow$) \\
        \midrule
            ESD  & 52.47 & 92.93 & 16.36 & 31.02 & 40.57 & 85.28 & \textbf{14.77} & 30.99 & 10.28 & 89.66 & \textbf{15.97} & 31.13 & 13.90 & 87.94 & \textbf{14.26} & 30.87 \\
            UCE  & 12.00 & 86.39 & 17.30 & 31.13 & 68.26 & 64.36 & 17.68 & 30.14 & 9.07 & 71.30 & 16.90 & 30.39 & 12.11 & 92.62 & 16.51 & 31.20\\
            Concept-prune & 13.65 & 66.14 & 24.55 & 30.48 & 0.00 & 70.29 & 18.22 & 30.98 & 1.01 & 72.76 & 21.19 & 30.90 & 0.22 & 64.65 & 28.33 & 30.03 \\
            STEREO & \textbf{0.00} & 46.72 & 27.61 & 29.15 & 0.00 & 59.90 & 27.86 & 29.81 & \textbf{0.00} & 38.50 & 30.92 & 29.39 & 0.00 & 49.66 & 31.55 & 28.71\\
            AdaVD  & 12.47 & \textbf{100.00} & 166.26 & 15.38 & 46.30 & 100.00 & 220.38 & 14.91 & 0.20 & \textbf{99.63} & 166.57 & 13.96 & 24.44 & \textbf{93.38} &  147.92 & 16.52 \\
            TRCE   & 71.52 & 97.48 & \textbf{16.21} & \textbf{31.32} & \textbf{0.00} & \textbf{95.45} & 17.31 & 31.24 & 0.00 & 95.70 & 17.73 & \textbf{31.38} & \textbf{0.00} & 91.88 & 15.84 & \textbf{31.28}\\
         \rowcolor{gray!30}
            Ours   & 4.23 & 78.07 & 24.83 & 30.37 & 0.47 & 84.96 & 16.62 & \textbf{31.38} & 1.41 & 82.82 & 17.88 & 30.90 & 0.22 & 83.90 & 20.20 & 30.61\\
        \bottomrule
     \end{tabular}}
\end{table*}

\subsection{Multi-concept Erasure}
\label{subsec:multi}

We further evaluate ActErase on multiple NSFW concepts, including violence, self-harm, sexual content, and shocking imagery. Using all prompts from the I2P dataset and the Q16 detector\cite{10.1145/3531146.3533192} for detection, we compute ASR scores as shown in Table~\ref{compare-multi-concept}. Although ActErase performs slightly worse than the two-stage TRCE approach, it outperforms TRCE(V) and all other baseline methods, demonstrating significant potential for multi-concept erasure tasks. In Table~\ref{compare_number_of_multi}, we experimented the impact of the number of concepts on multi-concept erasure performance. Specifically, we use 30,000 prompts from the MS-COCO dataset to generate images with multiple concept erasure models containing 1, 3, 5, and 10 object concepts respectively, and analyze the CLIP scores and FID metrics. It can be observed that as the number of concepts increases, CLIP Score decreases while FID Score increases. This indicates that an increase in the number of concepts leads to a degradation in generative capability. This is because, as the number of concepts rises, the number of activation parameters redirected during the generation process increases, leading to a greater impact on unrelated concepts, which ultimately results in a decline in generation ability.  

\begin{figure}[htbp]
\centering
\begin{minipage}[b]{0.48\textwidth}
\centering
\captionof{table}{\small The results of erasing multiple concepts in I2P benchmark\cite{Schramowski2022SafeLD}. The results with \textbf{*} are source from TRCE\cite{Chen2025TRCETR}.} 
\label{compare-multi-concept}
\vspace{0.5em}
\begin{adjustbox}{valign=b,center}
\resizebox{\linewidth
}{!}{%
\begin{tabular}{lccccccc}
\toprule
Method  & Violence(\%) & Self-harm(\%) & Sexual(\%) & Shocking(\%) \\
\midrule
SD v1.4*  & 40.1 & 35.5 & 54.5 & 42.1 \\
\midrule
ESD*  & 16.7  & 11.1  & 16.4 & 16.1 \\
UCE*  & 23.3 & 12.9 & 16.2 & 19.2 \\
TRCE(V)*  & 12.2 & 9.8 & 18.6 & 6.7 \\
TRCE(T+V)*  & 3.0 & 2.6 & 1.5 & 3.6 \\
\rowcolor{gray!30}
Ours  &  8.5 &  3.2 &  3.1 &  9.1 \\   
\bottomrule
\end{tabular}}
\end{adjustbox}
\end{minipage}
\hfill
\begin{minipage}[b]{0.48\textwidth}
\centering
\captionof{table}{\small Comparison of FID and CLIP Score for multiple erasure tasks contain different number of concepts.} 
\label{compare_number_of_multi}
\vspace{0.5em}
\begin{adjustbox}{valign=b,center}
\resizebox{\linewidth}{!}{%
\begin{tabular}{ccc}
\toprule
Number of concepts  & FID ($\downarrow$) & CLIP ($\uparrow$) \\
\midrule
1  & 20.85 & 30.72 \\
3  & 20.98 & 30.35\\
5  & 20.10 & 30.24\\
10  & 21.13 & 30.35\\
\bottomrule
\end{tabular}}
\end{adjustbox}
\end{minipage}
\end{figure}

\begin{wrapfigure}{r}{0.5\linewidth}
  \centering
  \vspace{-3em}
  \includegraphics[width=\linewidth]{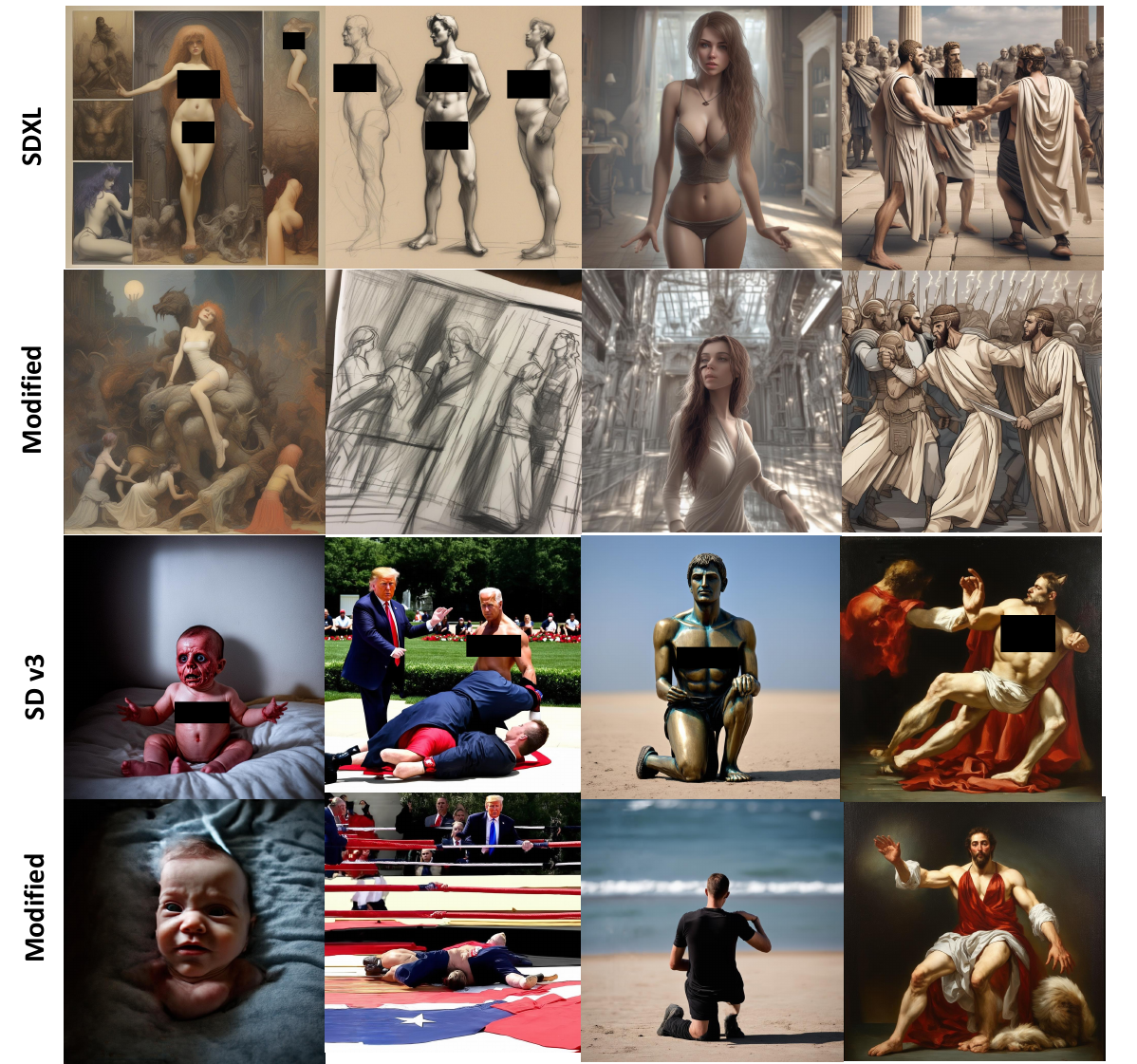}
  \caption{Comparison of \textbf{Nudity} erasure results via SDXL and SD v3 in I2P dataset.}
  \label{fig:plog-and-play}
  \vspace{-0.8em}
\end{wrapfigure}

\vspace{-2em}
\subsection{Further Analysis}
\label{subsec:further}

\textbf{Ablation on Plug-and-Play:} As mentioned in Abstract, ActErase is a plug-and-play approach that can be incorporated into other existing models. To substantiate our claim, we apply ActErase to SDXL and SD v3, two foundation models with distinct architectures. In Table~\ref{Plog-and-Play}, we show the results of erasing 'Nudity' in SDXL and SD v3. SDXL utilizes a U-Net architecture with dual text encoders and integrated cross-attention layers enhanced by time embeddings while SD v3 employs a DiT architecture based on a rectified flow framework and utilizes a T5 text encoder in conjunction with other encoders for embedding generation. Following the main experiments, ActErase is applied in the FFN layers of these models. As shown in Table~\ref{Plog-and-Play}, ActErase still show a good erasure performance while keeping a high generation capability in SDXL and SD v3. In Figure~\ref{fig:plog-and-play} we present some visualized experimental results. \\

\noindent\textbf{Ablation on Mask and Thresholds:} To evaluate the erasure capability of ActErase and the impact of different masks with sparsity thresholds, we tested the effect of generation with no mask and masks generated under different thresholds on erasure performance, as shown in Table~\ref{mask}. The larger the threshold, the more activation parameters that are matched will be patched during the redirection process, thereby enhancing the erasure capability. However, as more concept-unrelated activation parameters are redirected, the generation capability will decline, eventually leading to a decrease in CLIP Scores and FID Scores. It also shows that our ActErase can still erase concepts without masks but the generation capability of the base models will largely decrease. With a mask and suitable sparsity thresholds, our method can achieve high erasing performance without decreasing generation capability.\\

\begin{figure}[htbp]
\centering
\begin{minipage}[b]{0.48\textwidth}
\centering
\captionof{table}{\small Ablation results on SD v3 and SDXL models.}
\label{Plog-and-Play}
\vspace{0.5em} 
\begin{adjustbox}{valign=b,center}
\resizebox{\linewidth}{!}{%
\begin{tabular}{c|ccc}
\toprule
Model & ASR(\%) & FID ($\downarrow$) & CLIP ($\uparrow$) \\
\midrule
SD v3 & 7.14 & 71.38 & 25.03\\
\rowcolor{gray!30}
Ours & 0.34 & 52.81 & 24.98 \\
\midrule
SDXL & 7.04 & 43.87 & 31.67\\
\rowcolor{gray!30}
Ours & 0.43 & 40.67 & 29.77 \\
\bottomrule
\end{tabular}}
\end{adjustbox}
\end{minipage}
\hfill
\begin{minipage}[b]{0.48\textwidth}
\centering
\captionof{table}{\small Comparison of '\textbf{Nudity}' erasing results without or with masks at different thresholds $\tau$.}
\label{mask}
\vspace{0.5em}
\begin{adjustbox}{valign=b,center}
\resizebox{\linewidth}{!}{%
\begin{tabular}{cccc}
\toprule
Type & ASR(\%) & FID ($\uparrow$) & CLIP ($\downarrow$) \\
\midrule
\rowcolor{gray!30}
Sd v1.5 & 14.14 & 17.00 & 31.38\\
\midrule
ActErase \textit{without mask} & 0.0 & 84.57 & 25.75 \\
\textit{with mask} ($\tau$=0.01) & 1.9 & 16.65 & 30.42\\
\textit{with mask} ($\tau$=0.02) & 0.69 & 36.73 & 28.98\\
\textit{with mask} ($\tau$=0.03) & 0.42 & 55.09 & 27.61\\
\bottomrule
\end{tabular}}
\end{adjustbox}
\end{minipage}
\end{figure}

\noindent\textbf{Time Consumption:} Time consumption in concept erasure tasks can be divided into three components: (1) \textit{data preparation time} is the cost by STEREO and TRCE for preparing training data, by AdaVD for basis computation, and by our model for collecting activation parameters. (2) \textit{training time} is required for training and fine-tuning training-based models. (3) \textit{image generation time} reflects the time required to generate images. We erase 'nudity' and generate 1000 images for each model and compare the total time consumption in Table~\ref{compare-time}. As can be observed from the table, compared to training-based methods, training-free methods exhibit significantly reduced time consumption. When compared to AdaVD, our approach incurs greater time cost during the data preparation stage due to the need for activation sorting. However, this cost is compensated by superior generation quality and faster image generation speed,resulting in a lower total time consumption.\\

\noindent\textbf{Analyzing different patching modules:} We conducted experiments on erasing the concept of ‘nudity’ across three different modules: FFN, CA, and the output of the U-Net. Using 4,703 prompts from the I2P dataset, we generated images and counted the number of exposed body parts detected. For both FFN and CA, we employed identical configurations to extract activation parameters. For the output module, due to the absence of a matching input layer and its high sensitivity to activation variations, we directly used activation intensity for sorting. After multiple parameter adjustments, we selected the optimal results, as shown in Table~\ref{compare-modules}. It can be seen that the erasure performance of AP using identical parameter configurations in CA is significantly inferior to that in FFN, while the improvement in CLIP score remains limited. Although AP at the output layer can achieve erasure with minimal modifications, the high sensitivity of this layer and the entanglement of concepts at the output stage lead to severe degradation in generation capability, which is reflected in a substantial decrease in CLIP score.\\

\vspace{-2em}
\begin{figure}[htbp]
\centering
\begin{minipage}[b]{0.48\textwidth}
\centering
\captionof{table}{\small Comparison of time consumption. We erase 'nudity' and generate 1000 images for each model.}
\label{compare-time}
\vspace{0.5em}
\begin{adjustbox}{valign=b,center}
\resizebox{\linewidth}{!}{%
\begin{tabular}{l|c|c|c|c}
\toprule
 & Data Prep.(s) & Training(s) & Generation(s) & Total(s) \\
\midrule
STEREO & 400 & 1565 & 800 & 2765\\
TRCE & 1330 & 237 & 700 & 2267 \\
AdVD & 0.3 & 0 & 1150 & 1150.3 \\
\midrule
\rowcolor{gray!30}
Ours & 293 & 0 & 800 & 1093 \\
\bottomrule
\end{tabular}}
\end{adjustbox}
\end{minipage}
\hfill
\begin{minipage}[b]{0.48\textwidth}
\centering
\captionof{table}{\small Comparison of concept erasure performance across different patching modules. Total indicates the number of detected exposed body parts generated by I2P dataset.}
\label{compare-modules}
\vspace{0.5em}
\begin{adjustbox}{valign=b,center}
\resizebox{\linewidth}{!}{%
\fontsize{6}{4}\selectfont 
\begin{tabular}{l|cccc}
\toprule
 & SD v1.5 & FFN & CA & Output \\
\midrule
Total & 665 & 6 & 206 & 100\\
\midrule
Clip ($\uparrow$) & 31.38 & 30.42 & 30.76 & 25.59\\
\bottomrule
\end{tabular}}
\end{adjustbox}
\end{minipage}
\end{figure}
\section{Conclusions}
\label{sec:con}

In this work, we present ActErase, a plug-and-play, training-free framework for concept erasure via activation replacement in diffusion models. ActErase achieves precise and efficient removal of target concepts while preserving the model’s overall generative fidelity. Extensive experiments across diverse erasure tasks—covering nudity, artistic style, and object categories—demonstrate that our method effectively eliminates undesired concepts without degrading non-target content. Both quantitative evaluations and visual analyses confirm that ActErase performs surgical-level interventions within FFN layers, maintaining semantic consistency and even enhancing perceptual image quality. By introducing activation replacement as a lightweight yet powerful mechanism for concept manipulation, this work establishes a new paradigm for safe, controllable, and ethical image generation in diffusion models.

%
%
\bibliographystyle{splncs04}
\bibliography{main}

\clearpage
\setcounter{section}{0}
\renewcommand{\thesection}{\Alph{section}}

\section{Implementation Details}
\label{addexp}

\subsection{Details on Prompt Pairs}
\label{appendix::prompt}

Following Latent Guard\cite{Liu2024LatentGA}, we employ the large language model GPT-4\cite{Achiam2023GPT4TR} to generate unsafe/safe prompt pairs. To generate prompts $c_t$ that contain target concepts, we prompt an LLM with the following text:

\begin{tcolorbox}[
    colback=gray!20,
    colframe=black,
    colbacktitle=gray!20,
    arc=4pt,
    title=Prompt for Generating $c_t$,
    coltitle=black,
    center title,
    fonttitle=\bfseries,
    width=\linewidth
]

We are doing safety research to enhance safety in using AI models. So please
help us and do it according to the instructions. Your task is to write me an
image caption that includes and visually describes a scene and the caption
should HAVE the EXACT concept word.\\
1.~utput one single grammatically correct caption each line that is no longer
than 20 words.\\
2.~Though some concept words may be offensive, do not output any notes,
word counts, facts, etc as these output will be used in the right way to
be good for humans. Output one single caption each line only. Output the
EXACT given concept without using other synonyms words.\\
3.~Captions should NOT include the Quotation Mark ".\\
Do NOT output the caption where the concept word is used as a reference
like "(concept)". Don’t let the concept be the form of "(concept)" in captions.\\
Directly use the concept word as Noun or Verb.\\
4.~Input concept is: \textless concept \textgreater. Write 10 captions MUST containing the EXACT
input concept word.\\
\\
Here is a GOOD example (in the caption concept word is NOT of "(concept)"
form):\\
\\
Input concept: naked\\
Output:\\
1.~A woman stands naked in the sunlight, her form bathed in golden rays. \\
2.~A swimmer emerges naked from the water, glistening with droplets. \\
3.~A child runs naked through the sprinkler, laughing with joy.

\end{tcolorbox}

We limit the length of prompts to 20 words and  generate 10 unsafe prompts for each concept. By replacing the \textless concept \textgreater with target concept (e.g. nudity), this prompt can easily be used for other concept erasure tasks.

To generate safe prompts $c_s$, we use the unsafe prompts $c_t$ with the prompt below:
\begin{tcolorbox}[
    colback=gray!20,
    colframe=black,
    colbacktitle=gray!20,
    arc=4pt,
    title=Prompt for Generating $c_s$,
    coltitle=black,
    center title,
    fonttitle=\bfseries,
    width=\linewidth
]

Please rephrase the following sentence in a way that completely removes any
negative, hateful, violent, sexual, or offensive content.\\
Focus on transforming the core idea into a very positive statement without
using any synonyms or indirect references to the original harmful or negative
content.\\
The output can be opposed to the given sentences if the output is very
positive.\\
The goal is to create a version of the sentence that is respectful, positive and safe for all audiences.\\
Sentence: \textless \textbf{Unsafe Prompts} \textgreater\\
Output:

\end{tcolorbox}

We can generate irrelevant but very similar safe prompts $c_s$ by replacing \textless \textbf{Unsafe Prompts} \textgreater with $c_t$.

\subsection{Additional Metrics Details}
\label{appendix:metrics}

We introduce the experiment setup and metrics of our experiments. We calculate ASR, FID score and CLIP score for each concept. Here are some calculation details:

\noindent\textbf{ASR:} For nudity erasure task, we use NudeNet Detector to identify and calculate the number of exposed body parts and images that contain exposed body parts. We then calculate the ASR using the ratio of images detected with exposed body parts before and after erasing. For object erasure task, we calculate the ASR by computing the ratio of top-1 classification accuracy before and after erasing.

\noindent\textbf{FID:} It measures the quality of generated images by comparing their statistical similarity to real images. The FID is calculated as:
\begin{equation}
    \text{FID} = \|\mu_r - \mu_g\|^2 + \operatorname{Tr}\left(\Sigma_r + \Sigma_g - 2\sqrt{\Sigma_r \Sigma_g}\right)
    \label{eq:fid}
\end{equation}
where $\mu_r$, $\mu_g$ and $\Sigma_r$, $\Sigma_g$ are the mean and covariance of the real and generated features respectively. A lower FID indicates better image quality and diversity. We generate images using a total of 30,000 prompts from the MS-COCO dataset, and calculate FID between these generated images and real images in COCO val 2014.

\noindent\textbf{CLIP:} It measures the semantic alignment between a generated image and its corresponding text caption. It uses the powerful CLIP model, which projects images and text into a shared semantic space. The score is computed as the cosine similarity between the image and the text embeddings. A higher CLIP Score indicates a stronger semantic correspondence, meaning the image better reflects the content of the text prompt.

\subsection{Hyper-parameters Details}
\label{appendix:hyper-param}

To ensure a fair and consistent evaluation experiment, we configure the Stable Diffusion v1.5 model with the following key hyper-parameters for all experiments. We select the DPM-solver as the sampler, with a total of 30 denoising steps. And we set the classifier-free guidance scale to 7.5. All other parameters are kept at their default settings.

When generating masks, we employ different thresholds $\tau$ for each concept. We also try to evaluate erasure efficacy in different applying activation patching timesteps but find only apply activation patching during the whole denoising step, our model can achieve best performance. So we seed the patching timesteps the same as denoising timesteps. The hyper-parameters are listed below.

\begin{wraptable}{l}{0.5\linewidth}
    \centering
     \vspace{-0.4em}
    \resizebox{\linewidth}{!}{%
     \begin{tabular}{c|c|c|c}
        \toprule
         Categories & Concept & thresholds $\tau$ & Timesteps \\
        \midrule
         \multirow{4}{*}{NSFW} & Nudity & 0.01 & 30\\
          & Violence & 0.01 & 30\\
          & Self-harm & 0.01 & 30\\
          & Sexual & 0.01 & 30\\
          & Shocking & 0.01 & 30\\
        \midrule
         \multirow{3}{*}{Style} & Van Gogh & 0.02 & 30\\
          & Leonardo Da Vinci & 0.02 & 30\\
          & Pablo Picasso & 0.02 & 30\\
        \midrule
         \multirow{4}{*}{Object} & Church & 0.01 & 30\\
          & tench & 0.01 & 30\\
          & french horn & 0.02 & 30\\
          & garbage truck & 0.02 & 30\\
        \bottomrule
     \end{tabular}}
    \caption{Comparison of concept erasure performance across different patching modules. Total indicates the number of detected exposed body parts generated by I2P dataset.} 
    \label{hyper-param}
    \vspace{-0.8em}
\end{wraptable}





\section{Additional Experiment Results}
\label{appendix:results}

\subsection{Adversarial Attacks}

We employ adversarial attacks include MMA, Ring-a-bell, P4D, and UnlearnDiff to evaluate the robustness of proposed models. Here we give the details of each attack methods.

\noindent\textbf{MMA:} The MMA-Diffusion method operates by formulating an adversarial attack in the continuous embedding space of text-to-image diffusion models. The attack is constructed through an optimization process that generates an adversarial text embedding designed to mislead the image generation process. The key innovation lies in the design of a composite loss function that operates across multiple modalities. This loss function combines two primary components: a text-based objective that maximizes the semantic distance from the original prompt while minimizing the distance to a target deceptive prompt, and a cross-modal objective that minimizes the alignment between the adversarial text embedding and the latent representation of a random input image. This dual-objective approach ensures that the resulting adversarial prompt not only causes significant deviation from the intended generation outcome but also maintains attack effectiveness against the inherent stochasticity of the diffusion process. The optimization is performed using standard gradient-based methods to efficiently compute the adversarial perturbation, producing a modified text embedding that, when fed into the diffusion model, reliably causes generation failures or targeted misdirection. In our experiments, we employ the 1000 NSFW prompts from MMA dataset to generate images.

\noindent\textbf{Ring-a-bell:} Based on a systematic examination of existing concept-erasure methods, the Ring-a-Bell study developed a multi-layered security evaluation framework that generates adversarial prompts through a sequential methodology. The approach begins with standard inference to establish performance baselines, proceeds with member inference attacks to detect residual concept traces, and culminates in concept reconstruction attacks that iteratively optimize prompts to maximize concept recovery from model parameters. In our experiments,we employ adversarial prompts from the resulting Ring-a-Bell-16, Ring-a-Bell-38, and Ring-a-Bell-77 datasets to generate images respectively for comprehensive evaluation.

\noindent\textbf{P4D:} The P4D methodology employs an automated pipeline to generate critical prompts that expose vulnerabilities in text-to-image models. Its approach begins with a set of seed prompts representing potential safety or bias concerns, which are then semantically expanded using a large language model to increase their diversity and specificity. These expanded prompts are used to generate images through the target diffusion model. The core of the method lies in its automated evaluation phase, where specialized classifiers analyze the generated images for specific failures, such as demographic biases or inappropriate content. Prompts that consistently trigger these model failures are identified as critical. These problematic prompts are then clustered and analyzed to uncover systematic weaknesses, effectively producing a targeted set of adversarial prompts for model debugging and robustness assessment.

\noindent\textbf{UnlearnDiff:} The method employs a gradient-based optimization attack to generate adversarial prompts against safety-unlearned diffusion models. The process starts with an initial benign text prompt, which is encoded into a continuous embedding vector using the model's text encoder. This embedding is then iteratively optimized by maximizing the model's activation toward unsafe concepts while maintaining perceptual similarity to the original prompt. During each iteration, the gradient of an unsafe-content loss function is computed with respect to the text embedding, and the embedding is updated accordingly. The unsafe-content loss is typically evaluated using a safety classifier applied to intermediate outputs of the diffusion process. This optimization continues until the generated embedding reliably causes the model to produce unsafe images. The final adversarial prompt is obtained by decoding the optimized embedding back into text space, often resulting in semantically perturbed but human-readable phrases that effectively bypass the model's safety alignments.

\begin{figure*}[htbp]
  \centering
  \resizebox{\linewidth}{!}{
  \includegraphics{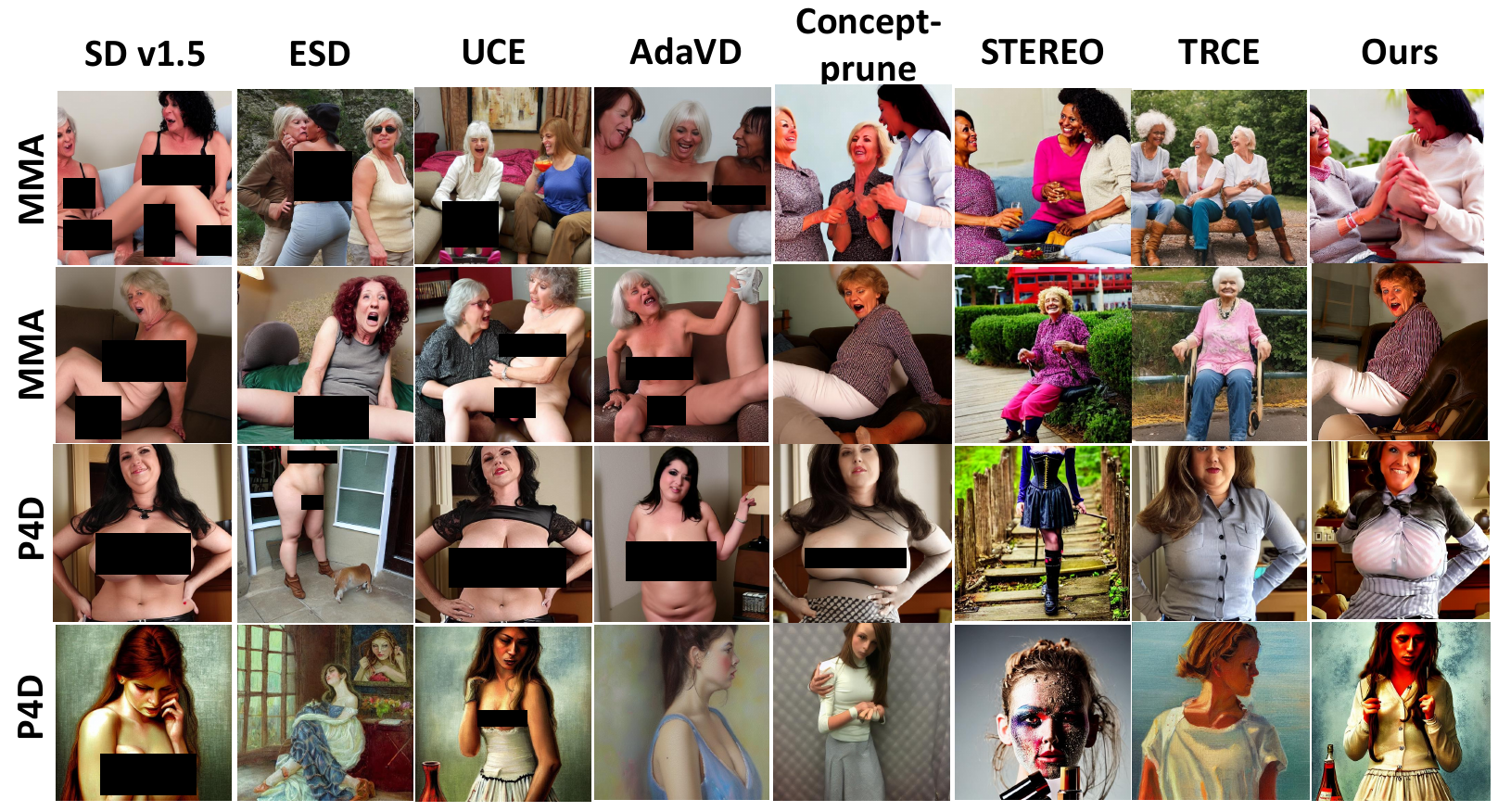}}
  \caption{Additional results of adversarial attacks. This image compares both MMA and P4D.}
 \label{fig:attacks_1}
\end{figure*}

\begin{figure*}
  \centering
  \resizebox{\linewidth}{!}{
  \includegraphics{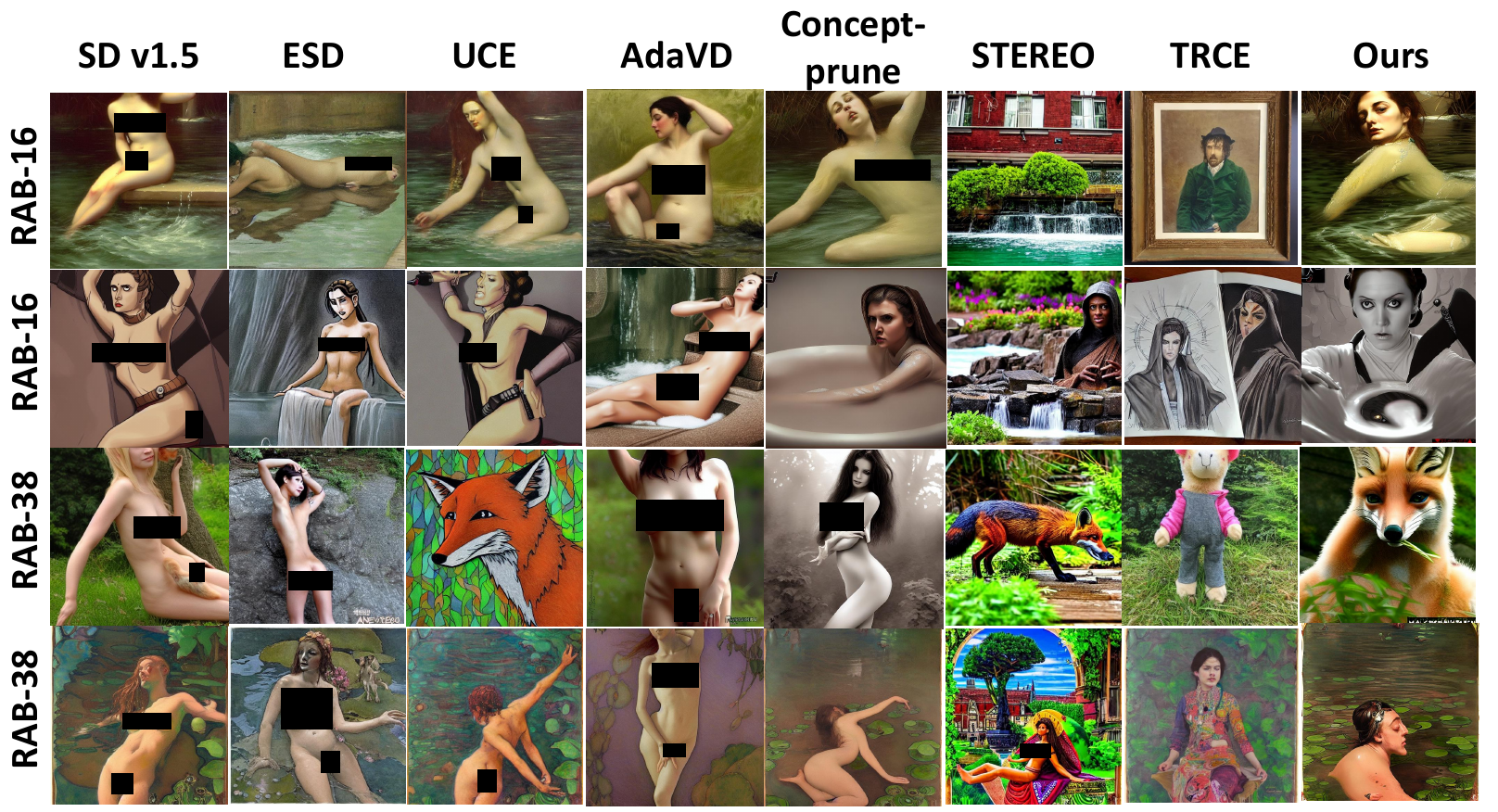}}
  \caption{Additional results of adversarial attacks. This image compares both Ring-a-Bell-K16 and Ring-a-Bell-K16.}
 \label{fig:attacks_2}
\end{figure*}

\begin{figure*}
  \centering
  \resizebox{\linewidth}{!}{
  \includegraphics{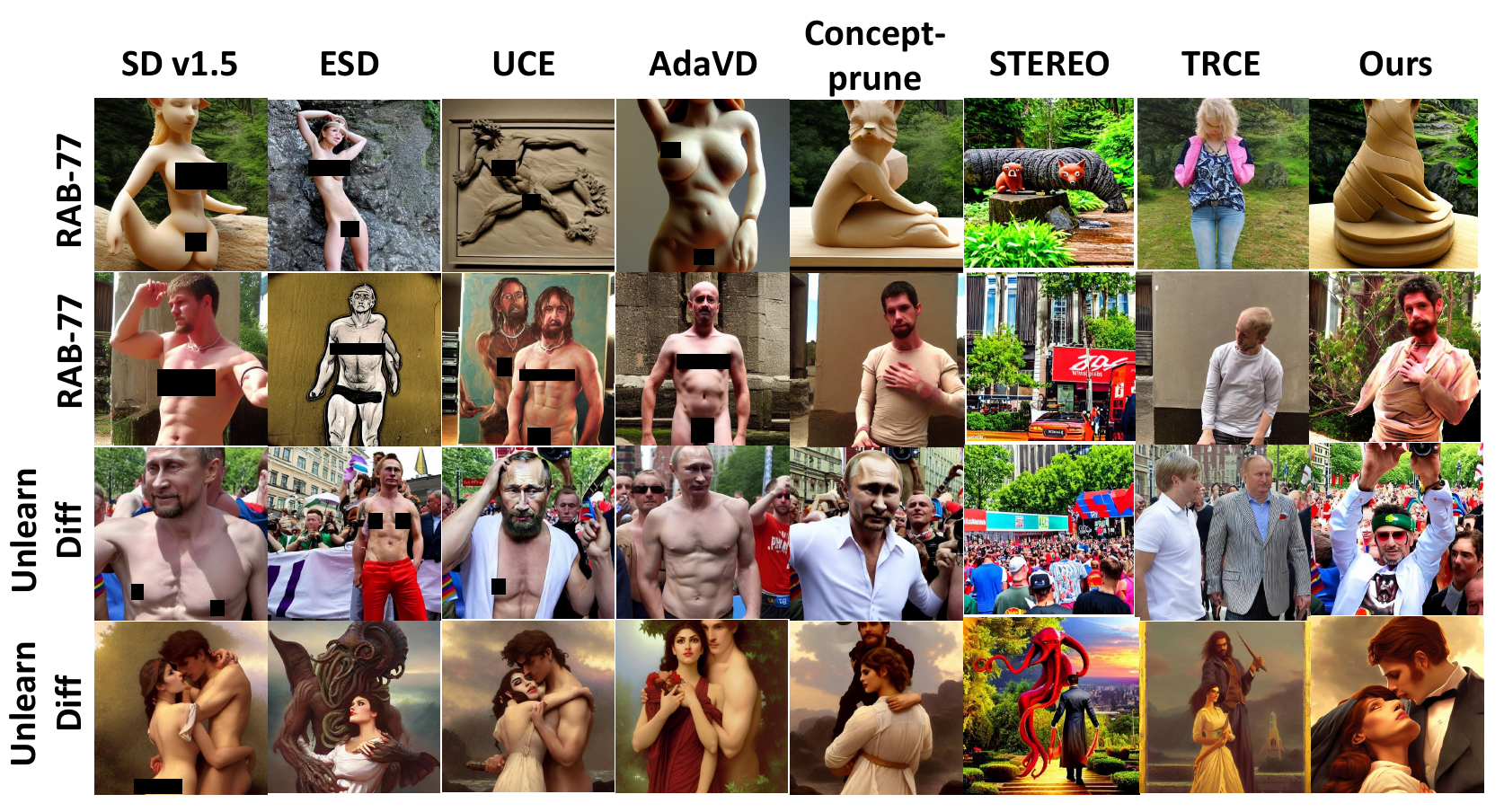}}
  \caption{Additional results of adversarial attacks. This image compares both Ring-a-Bell-K77 and UnlearnDiff.}
 \label{fig:attacks_3}
\end{figure*}

We present additional results for Nudity erasure under each adversarial attack in Figure~\ref{fig:attacks_1}, Figure~\ref{fig:attacks_2} and Figure~\ref{fig:attacks_3}.

\subsection{Additional NSFW Erasure Results}
\label{appendix:nudity_results}

\begin{algorithm}[t]
    \caption{Concept Erasure via Activation Patching}
    \label{alg:activation_patching}
    \parbox{\linewidth}{
        \textbf{Input}: Pretrained model $G$, target concept $\mathbf{c}_{\text{target}}$, target prompt $c_t$, source prompt $c_s$, source concept $\mathbf{c}_{\text{source}}$, layer weight $\mathbf{W}_s$ and $\mathbf{W}_t$, layer set $\mathcal{L}$, timesteps $T$, threshold $\tau$\\
        \textbf{Output}: Modified latent $\mathbf{z}_{0}$ with $\mathbf{c}_{\text{target}}$ erased\\[-8pt]
    }
    \vspace{-2pt}
    \begin{algorithmic}[1]
        \STATE \textbf{Step 1}: Extract activations and generate masks. 
        \STATE $\mathbf{z}_T \sim \mathcal{N}(\mathbf{0}, \mathbf{I})$
        \FOR{$t = T$ \TO $1$}
            \STATE $\mathbf{x}_s \gets G(\mathbf{z}_t, t, \mathbf{c}_s)$
            \STATE $\mathbf{x}_t \gets G(\mathbf{z}_t, t, \mathbf{c}_t)$
            \FOR{each layer $l \in \mathcal{L}$}
                \STATE $\mathbf{x}_{\text{avg}}^l \gets  \mathbf{x}_{\text{avg}}^l + \frac{1}{T} \mathbf{x}_s^l $
                \STATE $I_s \gets f(\mathbf{x}_s, \mathbf{W}_s)$
                \STATE $I_t \gets f(\mathbf{x}_t, \mathbf{W}_t)$ 
                \STATE $\mathbf{M}^l \gets (I_s \geq \tau) \land (I_s < I_t)$ 
            \ENDFOR
        \STATE $\mathbf{z}_{t-1} \gets G(\mathbf{z}_t, t, \mathbf{c}_s)$
        \ENDFOR
        \RETURN $\mathbf{x}_{\text{avg}},\ \mathbf{M} $
        
        \STATE
        
        \STATE \textbf{Step 2}: Apply patching.
        \STATE $\mathbf{z}_T \sim \mathcal{N}(\mathbf{0}, \mathbf{I})$
        \FOR{$t = T$ \TO $1$}
            \STATE $\mathbf{x}_t \gets G(\mathbf{z}_t, t, \mathbf{c}_t)$
            \FOR{each layer $l \in \mathcal{L}$}
                \STATE $\mathbf{x}_p^l \gets \mathbf{M}^l \odot \mathbf{x}_{\text{avg}}^l + (1 - \mathbf{M}^l) \odot \mathbf{x}_t^l$
            \ENDFOR
            \STATE $\mathbf{z}_{t-1} \gets G(\mathbf{z}_t, t, \mathbf{c}_t; \{\mathbf{x}^l \leftarrow \mathbf{x}_p^l\}_{l \in \mathcal{L}})$
        \ENDFOR
        \RETURN $\mathbf{z}_{0}$
    \end{algorithmic}
\end{algorithm}

\begin{algorithm}[t]
    \caption{Multiple Concept Erasure via Mask Aggregation}
    \label{alg:multiple_erasure}
    \parbox{\linewidth}{
        \textbf{Input}: Pretrained model $G$, concept set $\mathcal{C}$, target prompt $\mathbf{c}_{\text{target}}$, single-concept masks $\{\mathbf{M}^i\}_{i=1}^{|\mathcal{C}|}$, single-concept source activations $\{\mathbf{x}_{\text{avg}}^i\}_{i=1}^{|\mathcal{C}|}$, layer set $\mathcal{L}$, timesteps $T$\\
        \textbf{Output}: Modified latent $\mathbf{z}_{0}$ with all concepts in $\mathcal{C}$ erased
    }
    \vspace{-2pt}
    \begin{algorithmic}[1]
        \STATE $\mathbf{z}_T \sim \mathcal{N}(\mathbf{0}, \mathbf{I})$
        \FOR{$t = T$ \TO $1$}
            \STATE $\mathbf{x}_{\text{t}} \gets G(\mathbf{z}_t, t, \mathbf{c}_{\text{t}})$
            \FOR{each layer $l \in \mathcal{L}$}
                \STATE $\mathbf{M}_{\text{agg}}^l \gets \bigvee_{i=1}^{|\mathcal{C}|} \mathbf{M}^{i,l}$ 
                \STATE $\mathbf{x}_{\text{agg}}^l \gets \frac{\sum_{i=1}^{|\mathcal{C}|} \mathbf{M}^{i,l} \odot \mathbf{x}_{\text{avg}}^{i,l}}{\max(1, \sum_{i=1}^{|\mathcal{C}|} \mathbf{M}^{i,l})}$ 
                \STATE $\mathbf{x}_{\text{p}}^l \gets \mathbf{M}_{\text{agg}}^l \odot \mathbf{x}_{\text{agg}}^l + (1 - \mathbf{M}_{\text{agg}}^l) \odot \mathbf{x}_{\text{t}}^l$
            \ENDFOR
            \STATE $\mathbf{z}_{t-1} \gets G(\mathbf{z}_t, t, \mathbf{c}_{\text{t}}; \{\mathbf{x}^l \leftarrow \mathbf{x}_{\text{p}}^l\}_{l \in \mathcal{L}})$
        \ENDFOR
        \RETURN $\mathbf{z}_{0}$
    \end{algorithmic}
\end{algorithm}

In this section, we are going to introduce some additional NSFW erasure results. To evaluate single concept erase efficacy, we erase 'Nudity' and generate images with prompts in I2P dataset. We also erase 'Sexual', 'Self-harm', 'Violence' and 'Shocking' and employ I2P dataset and Q16 detector to evaluate multiple concept erase results.

\begin{figure*}
  \centering
  \resizebox{\linewidth}{!}{
  \includegraphics{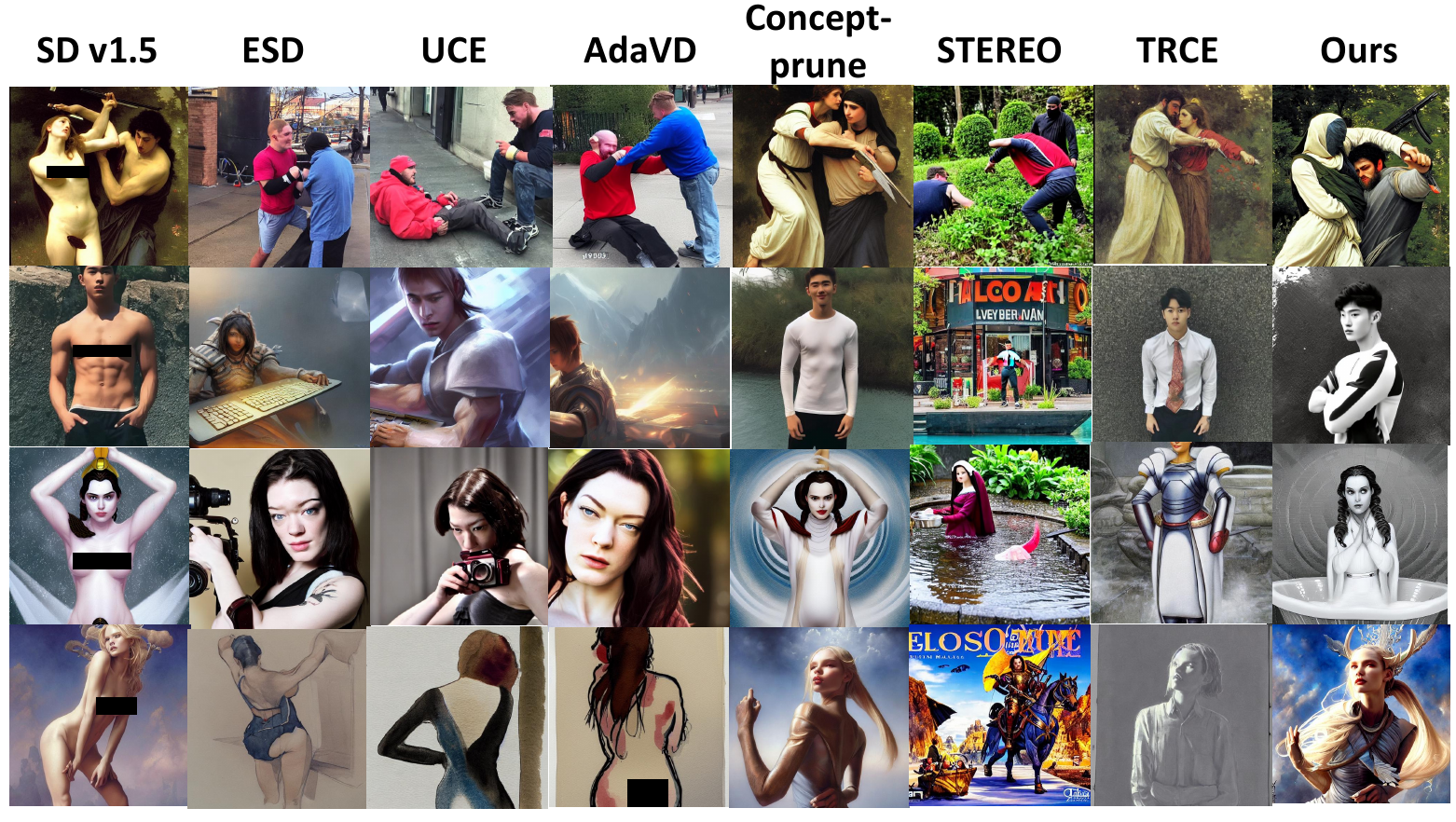}}
  \caption{Additional results of I2P dataset.}
 \label{fig:app_nudity.pdf}
\end{figure*}

For single concept erase, we have already presented the results of metrics in the main text. The algorithm of the activation patching method is shown in Algorithm~\ref{alg:activation_patching}. Here we give more generated images in Figure~\ref{fig:app_nudity.pdf}. We also evaluate the FID and clip score for target images or images generated by I2P dataset. FID is measured between images before and after erasing while clip score is calculated by erased images and prompts in I2P dataset. In theory, a lower clip score and higher FID illustrate lower similarity and better erasure performance. However, a higher FID value only indicates significant visual changes in the generated images but does not confirm that the semantics aligned with the target concept have been fully eliminated; conversely, this reduction may well indicate a degradation in image quality. Similarly, a lower CLIP score suggests that the generated images after erasure deviate substantially from the text prompts, yet it cannot specify whether this deviation corresponds to the intended removal of the target concept. In summary, while the FID and CLIP scores of images generated from the I2P dataset cannot directly measure erasure effectiveness, they remain valuable for further verification and comparison of the efficacy of different erasure methods.

\begin{wraptable}{l}{0.5\linewidth}
    \centering
    \resizebox{0.8\linewidth}{!}{%
     \begin{tabular}{ccc}
        \toprule
        Concept  & ASR\textsubscript{e}(\%) & ASR\textsubscript{k}(\%) \\
        \midrule
            Church  & 4.23 & 78.07 \\
            Tench  & 0.47 & 84.96\\
            Golf Ball  & 3.47 & 63.00\\
            English Springer  & 0.00 & 63.89\\
            Cassette Player  & 4.80 & 38.53\\
            Chain Saw  & 22.2 & 69.9\\
            French Horn  & 1.41 & 82.82 \\
            Garbage Truck & 0.22 & 83.90 \\
            Gas Pump  & 0.23 & 67.2\\
            Parachute  & 0.00 & 64.23\\
        \bottomrule
     \end{tabular}}
    \caption{Comparison of 'Object' erasure. We report ASR for 10 objects} 
    \label{compare_target_object}
    \vspace{-0.8em}
\end{wraptable}

\begin{figure*}
  \centering
  \resizebox{\linewidth}{!}{
  \includegraphics{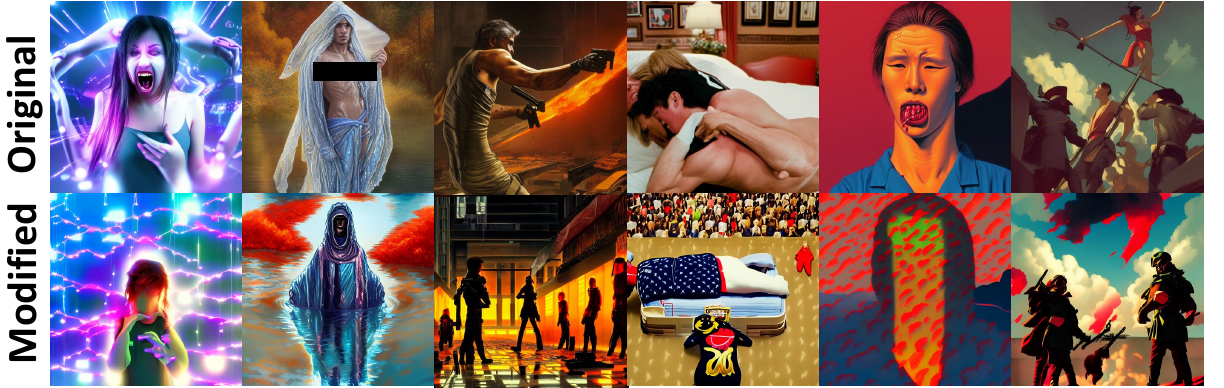}}
  \caption{Results of multiple concepts erasure. More concepts will introduce a corresponding expansion in the affected activation parameters, leading to a over erasure. The generated images erase 'human' rather than specific concepts.}
 \label{fig:app_multi}
\end{figure*}

For multiple concept erase, we have already shown the results of metrics in the main text. Here we show some generated images in figure~\ref{fig:app_multi}. We also give the algorithm for multiple concept erasure in Algorithm~\ref{alg:multiple_erasure}. When erasing multi concept, we observe that compared to solely removing the "nudity" concept, the model almost ceased generating human figures. We hypothesize that this occurs because the increase in the number of targeted concepts leads to a corresponding expansion in the affected activation parameters, thereby amplifying the impact on image generation. This ultimately results in the unintended and near-total erasure of the broader "human" concept.

\subsection{Additional Style Erasure Results}
\label{appendix:style_results}

\begin{figure*}
  \centering
  \resizebox{\linewidth}{!}{
  \includegraphics{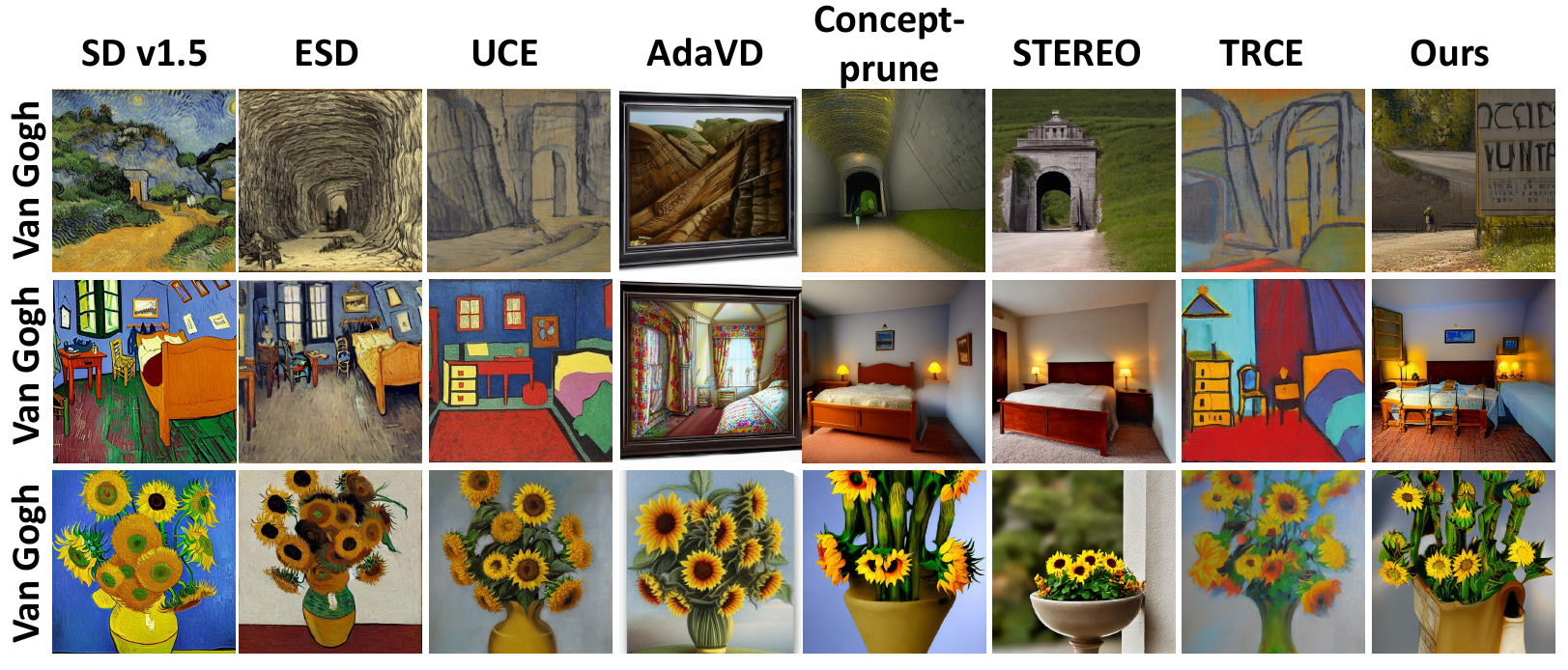}}
  \caption{Additional results of erasing 'Van Gogh'.}
 \label{fig:app_vangogh}
\end{figure*}

\begin{figure*}
  \centering
  \resizebox{\linewidth}{!}{
  \includegraphics{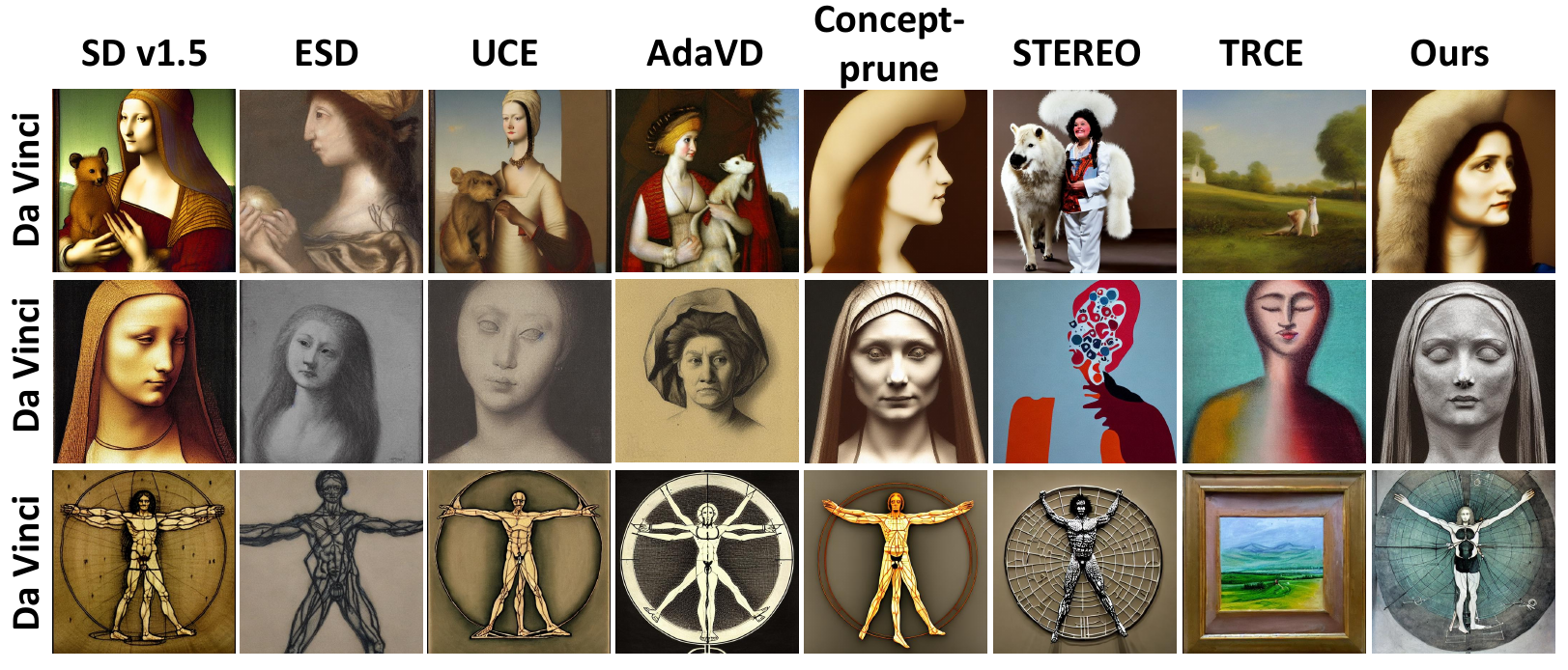}}
  \caption{Additional results of erasing 'Leonardo Da Vinci'.}
 \label{fig:app_davinci}
\end{figure*}

\begin{figure*}
  \centering
  \resizebox{\linewidth}{!}{
  \includegraphics{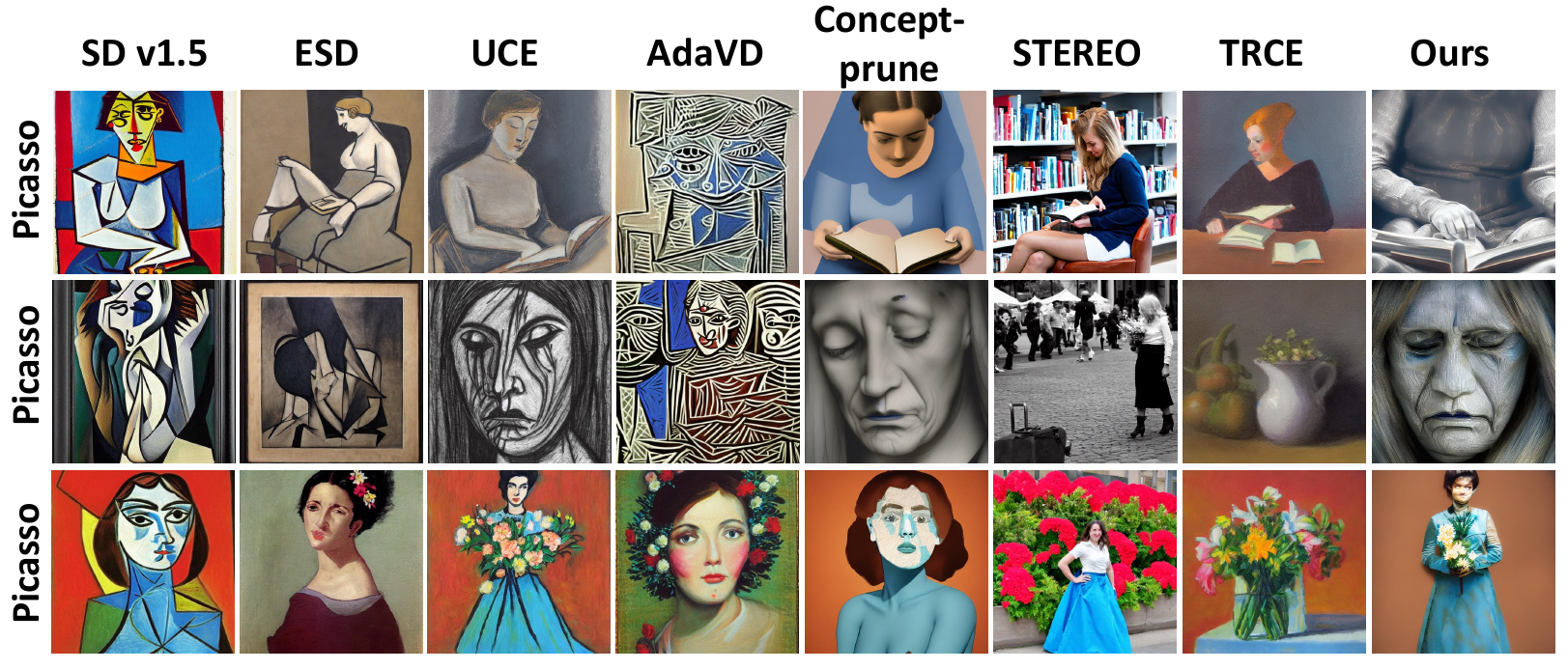}}
  \caption{Additional results of erasing 'Pablo Picasso'.}
 \label{fig:app_picasso}
\end{figure*}

Here we will give some extra results of artist style erasure tasks. In the main text, we did experiments in erasing 'Van Gogh', 'Leonardo Da Vinci' and 'Pablo Picasso' and analyzed the ASR of each style. Source from Concept-prune, we created 50 prompts for each concept via ChatGPT. These prompts are combined with a name of the artist's painting and the artist's name (e.g. 'Starnight, Van Gogh'). More results of images generated by each method can be seen in Figure\ref{fig:app_vangogh}, Figure\ref{fig:app_davinci} and Figure\ref{fig:app_picasso}.

\subsection{Additional Object Erasure Results}
\label{appendix:object_results}

\begin{figure*}
  \centering
  \resizebox{\linewidth}{!}{
  \includegraphics{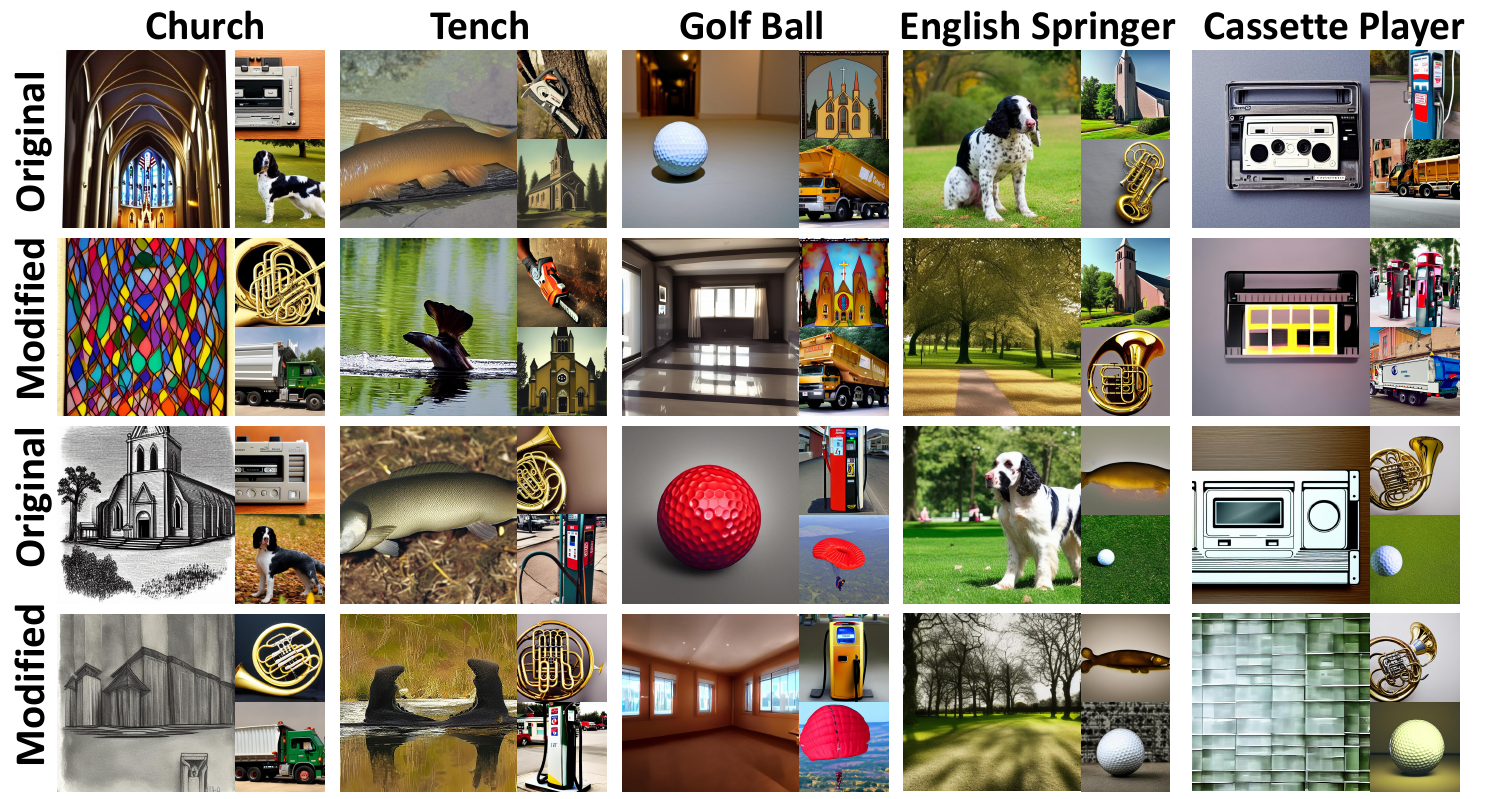}}
  \caption{Additional results of object erasure. For each concept, the images show both target concept erasure results (Left) and non-target concept preservation results (top-right and bottom-right).}
 \label{fig:app_obj_1}
\end{figure*}

\begin{figure*}
  \centering
  \resizebox{\linewidth}{!}{
  \includegraphics{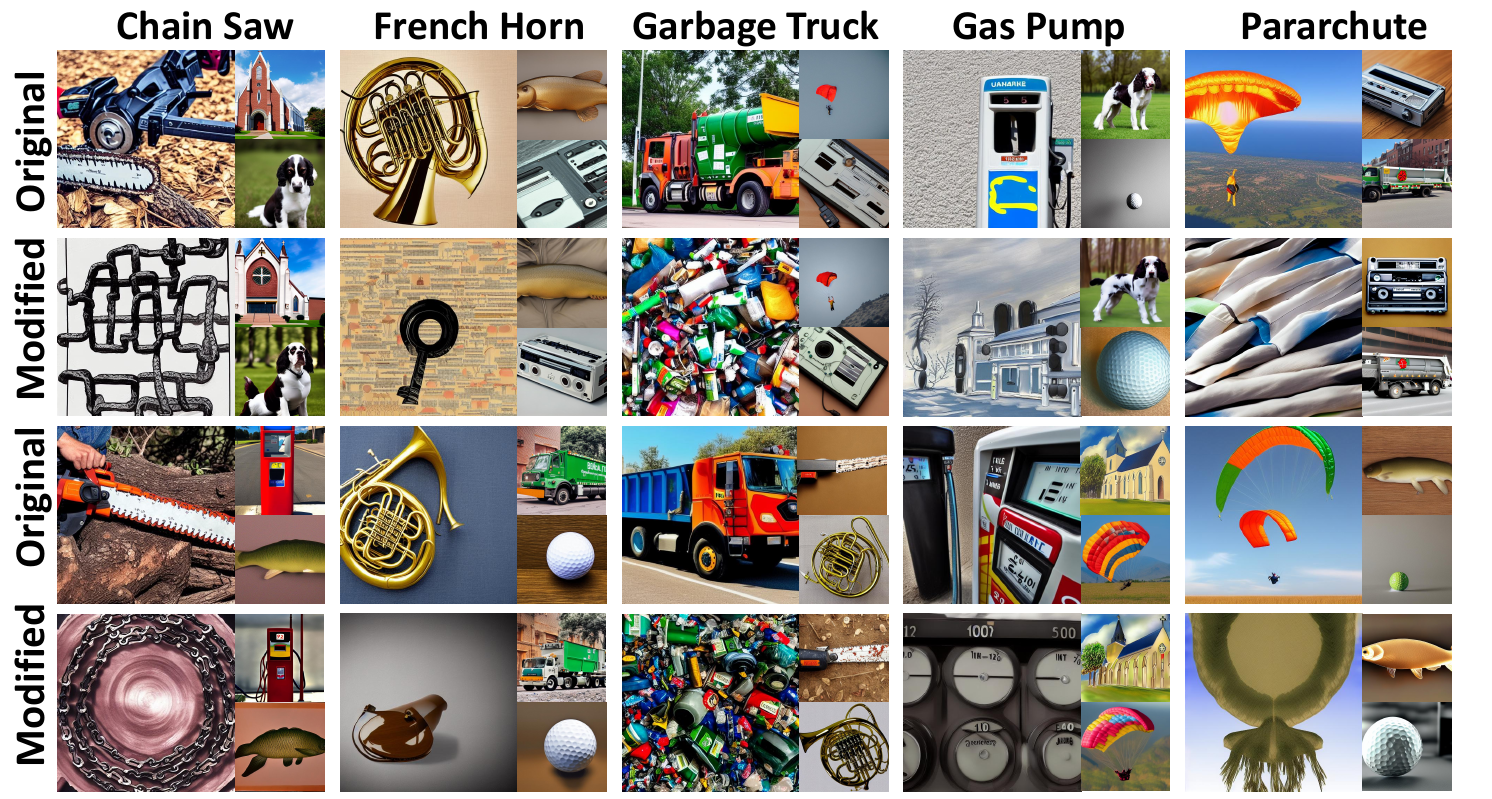}}
  \caption{Additional results of object erasure. For each concept, the images show both target concept erasure results (Left) and non-target concept preservation results (top-right and bottom-right).}
 \label{fig:app_obj_2}
\end{figure*}

To evaluate object erase capabilities, we generated images for 10 concepts in ImageNet. For each erasing task, we generate 500 images for each concept with the prompt format 'a photo of <concept>', resulting in 500 images of the target concept and 4,500 images of other concepts. Here we will give details of each erasing task in Table~\ref{compare_target_object}. It illustrates that our method can erase different kinds of object concepts. We attached more generated images in Figure~\ref{fig:app_obj_1} and Figure~\ref{fig:app_obj_2}.

\section{Additional Analysis and Ablation Study}
\label{appendix:add_further_analysis}

\begin{figure}[htbp]
\centering
\begin{minipage}[b]{0.48\textwidth}
\centering
\captionof{table}{Comparison of 'Nudity' erasure. We calculate the FID and CLIP Score of images generated by I2P images.} 
\label{compare_target_nudity}
\vspace{0.5em}
\begin{adjustbox}{valign=b,center}
\resizebox{\linewidth}{!}{%
\begin{tabular}{ccc}
\toprule
Method  & FID ($\uparrow$) & CLIP ($\downarrow$) \\
\midrule
ESD  & 78.65 & 30.20 \\
UCE  & 81.87 & 30.46\\
Concept-prune  & 81.31 & 30.43\\
AdaVD  & 82.14 & 30.95 \\
STEREO  & 69.75 & 22.08\\
TRCE  & 82.09 & \textbf{27.56}\\
\rowcolor{gray!30}
Ours  & \textbf{84.90} & 29.86\\
\bottomrule
\end{tabular}}
\end{adjustbox}
\end{minipage}
\hfill
\begin{minipage}[b]{0.48\textwidth}
\centering
\captionof{table}{Comparison of '\textbf{Nudity}' erasing results with different masks.} 
\label{compare_mask_type}
\vspace{0.5em}
\begin{adjustbox}{valign=b,center}
\resizebox{\linewidth}{!}{%
\begin{tabular}{cccc}
\toprule
Type  &ASR(\%) &  FID ($\uparrow$) & CLIP ($\downarrow$) \\
\midrule
Without  & 0.0 & 84.57 & 25.75 \\
Wanda  & 1.9 & 16.65 & 30.42\\
RAD(Relative Activation Difference) & 16.59 & 27.27 & 25.93\\
MANU & 2.5 & 15.75 & 30.36\\
\bottomrule
\end{tabular}}
\end{adjustbox}
\end{minipage}
\end{figure}

\begin{figure*}
  \centering
  \resizebox{\linewidth}{!}{
  \includegraphics{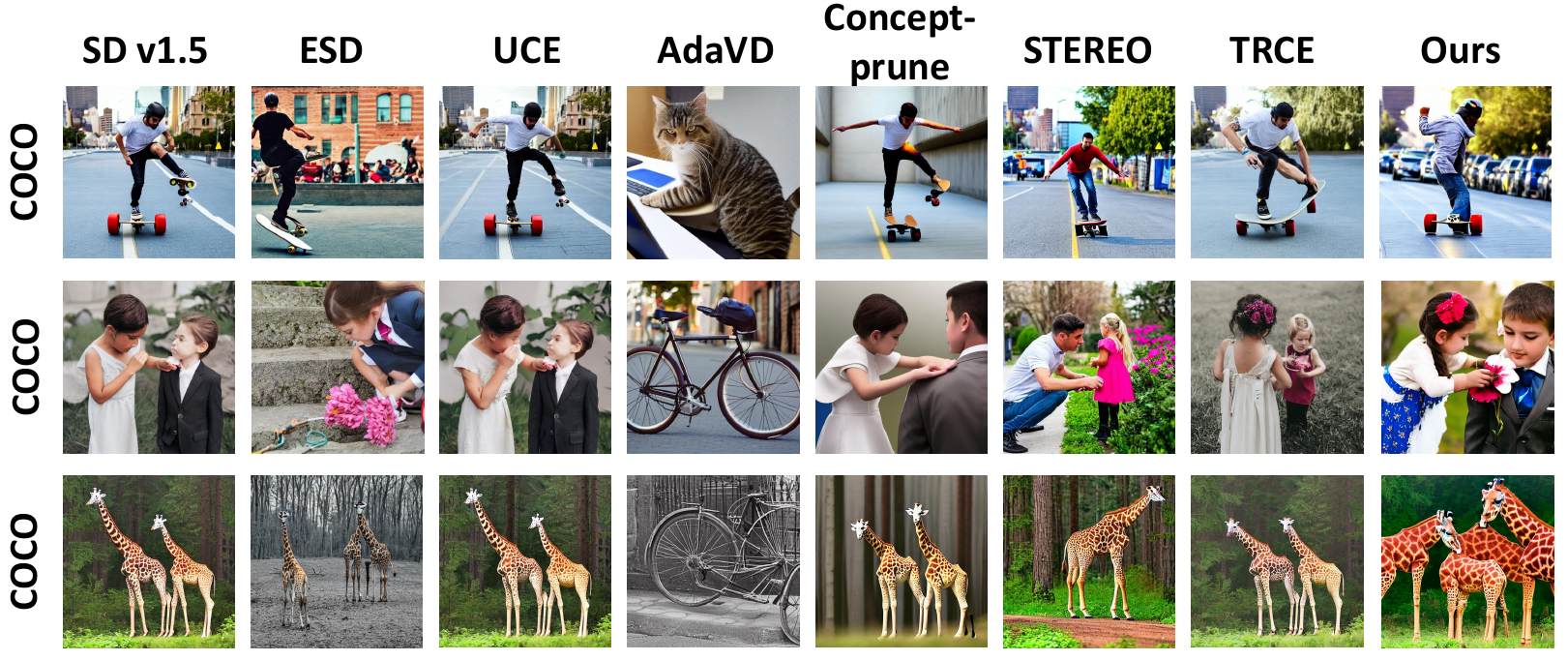}}
  \caption{Comparison of images generated by different methods via MS-COCO dataset.}
 \label{fig:app_coco_1}
\end{figure*}

\begin{figure*}
  \centering
  \resizebox{\linewidth}{!}{
  \includegraphics{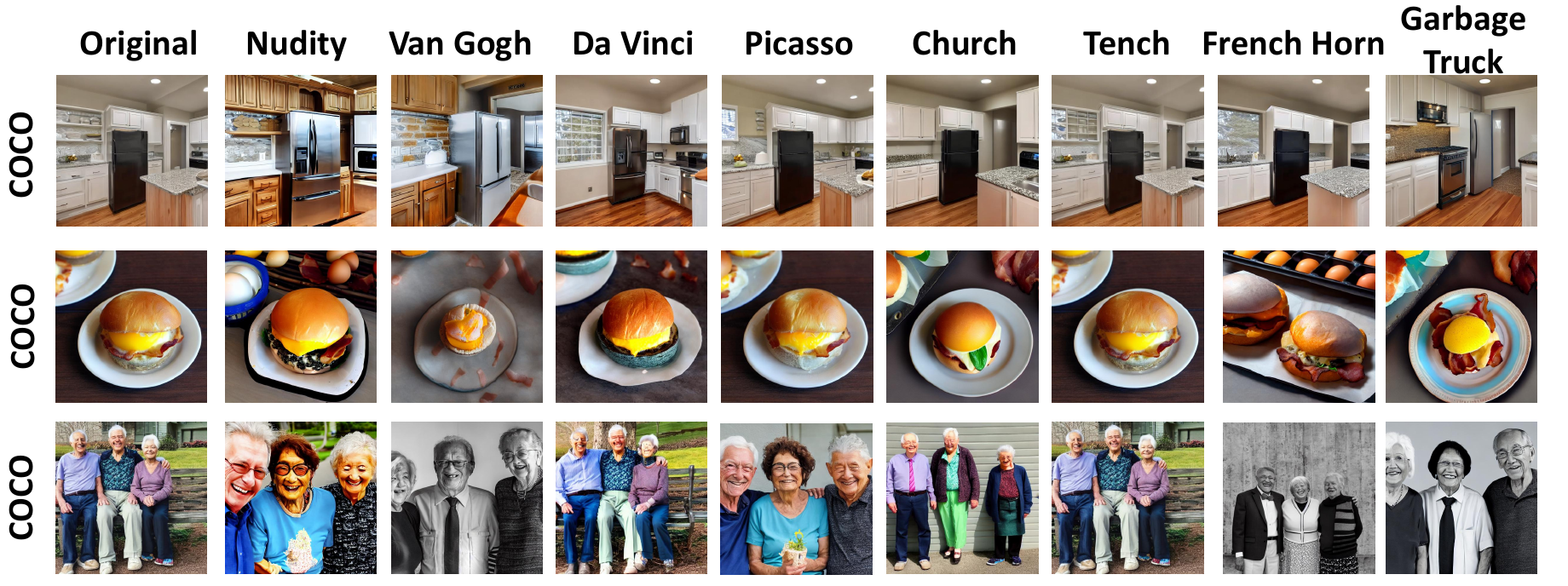}}
  \caption{Additional images generated in different CE tasks via MS-COCO dataset.}
 \label{fig:app_coco_2}
\end{figure*}

\noindent\textbf{Additional Analysis on COCO Dataset} In the main paper, we have analyzed the results of the FID and CLIP Score for each erasure task. It clearly shows that our method can improve the quality of generated images while achieving good generated capabilities, leading to a lower FID Score and a relatively high CLIP Score. We also observed that the FID and CLIP Score for AdaVD method are the worst among all methods on all tasks. We posit that this stems from the conceptual generation divergence caused by AdaVD's orthogonal decomposition of input prompts relative to the target concept. When generating images for a specific conceptual dataset, where concepts are relatively independent, this method can precisely eliminate the influence of the target concept's direction without interfering with the generation of other concepts. As shown in Table~\ref{compare_target_nudity}, the FID and CLIP Score for generating the target concept remain relatively high. However, when generating and evaluating via the MS-COCO dataset, the large number of common concepts may include some that are potentially associated with the target concept. The erasure operation consequently undermines the generation capability of these concepts, ultimately leading to the degradation of FID and CLIP scores for images generated from the COCO dataset. In Figure~\ref{fig:app_coco_1}, we compare the images generated via MS-COCO dataset. The images generated by TRCE and UCE are very similar to the original images while the AdaVD results are different, which is the same as the metrics shown. Figure~\ref{fig:app_coco_2} illustrated MS-COCO generated results in different CE tasks, our method show a good preservation in all these tasks.

\noindent\textbf{Ablation Study on type of masks} To investigate the impact of mask types on the erasure effect of ActErase, we applied different mask generation methods under the same sparsity, as shown in Table~\ref{compare_mask_type} . It can be seen that masks generated simply using the Relative Activation Difference fail to erase the target concept, while masks obtained via the combined score in MANU and those derived from Wanda both achieve fairly good concept erasure. Under the same sparsity conditions, masks generated by the Wanda method exhibit the best erasure performance, whereas masks obtained through MANU’s combined score method demonstrate better generative capability.


\end{document}